\pdfoutput=1

\documentclass[11pt]{article}

\usepackage{acl}

\usepackage{amsmath}
\usepackage{caption}
\usepackage{times}
\usepackage{latexsym}
\usepackage[T1]{fontenc}

\usepackage[utf8]{inputenc}

\usepackage{microtype}

\usepackage{inconsolata}

\usepackage{graphicx}
\usepackage{booktabs}
\usepackage{hyperref}
\usepackage{enumitem}
\usepackage{multirow}
\usepackage{multicol}
\usepackage{amssymb}
\usepackage{array}
\usepackage{subcaption}
\usepackage{xurl}

\usepackage{titletoc}

\newcommand{\dataReal}{ICDS$_R$}

\newcommand{\dataGenGemini}{ICDS$_G^g$}
\newcommand{\dataGenLlama}{ICDS$_G^l$}
\newcommand{\ncOld}{n2c2-2006}
\newcommand{\ncNew}{n2c2-2014}

\newcommand{\model}{PI-RoBERTa}

\title{Generation and De-Identification of Indian Clinical Discharge Summaries using LLMs}

\author{Sanjeet Singh\footnotemark[3]\footnotemark[1] \qquad Shreya Gupta\footnotemark[2]\thanks{Equal Contribution} \qquad Niralee Gupta\footnotemark[2] \\
\textbf{Naimish Sharma}\footnotemark[2] \qquad \textbf{Lokesh Srivastava}\footnotemark[2] \qquad \textbf{Vibhu Agarwal}\footnotemark[2] \\ \textbf{Ashutosh Modi}\footnotemark[3] \\
\footnotemark[3]Indian Institute of Technology Kanpur (IIT Kanpur) \qquad \footnotemark[2]Miimansa\\
\texttt{\{sanjeet, ashutoshm\}@cse.iitk.ac.in} \\ \texttt{\{shreya.gupta,niralee.gupta,naimish.sharma\}@miimansa.com} \\
\texttt{\{lokesh.srivastava,vibhu\}@miimansa.com}
}

\begin{document}
\maketitle

\begin{abstract}
The consequences of a healthcare data breach can be devastating for the patients, providers, and payers. The average financial impact of a data breach in recent months has been estimated to be close to USD 10 million. This is especially significant for healthcare organizations in India that are managing rapid digitization while still establishing data governance procedures that align with the letter and spirit of the law. Computer-based systems for de-identification of personal information are vulnerable to data drift, often rendering them ineffective in cross-institution settings. Therefore, a rigorous assessment of existing de-identification against local health datasets is imperative to support the safe adoption of digital health initiatives in India. Using a small set of de-identified patient discharge summaries provided by an Indian healthcare institution, in this paper, we report the nominal performance of de-identification algorithms (based on language models) trained on publicly available non-Indian datasets, pointing towards a lack of cross-institutional generalization. Similarly, experimentation with off-the-shelf de-identification systems reveals potential risks associated with the approach. To overcome data scarcity, we explore generating synthetic clinical reports (using publicly available and Indian summaries) by performing in-context learning over Large Language Models (LLMs). Our experiments demonstrate the use of generated reports as an effective strategy for creating high-performing de-identification systems with good generalization capabilities.
\end{abstract}


\section{Introduction} \label{sec:intro}

Over 330 million patient records in India have already been linked with a unique central ID \cite{pib_press_release}. To put this in perspective, the number roughly equals the total population of the United States. 
Several federal initiatives aimed at establishing standards for medical information exchange, adoption of controlled terminologies, and promoting open architecture-based systems for the management of patient records have seen a steady rise in the adoption of electronic health records within Indian healthcare institutions \cite{nrces,srivastava2016adoption}. This data represents an under-utilized resource that has profound implications for informing public policy, medical research and patient care. At the same time, it also poses some serious challenges. The risks of revealing patient identity even from data that has been anonymized are well known \cite{sweeney2013matching}. 
Privacy regulations such as GDPR 2016 \cite{gdpr2016} and the HIPAA Privacy Rule 2003 \cite{hipaa2003} lay down heavy penalties on non compliance with data safety protocols. A robust data de-identification pipeline is vital if we aim to unlock insights from these electronic patient histories.

\noindent Natural Language Processing (NLP) methods for de-identification are known to perform significantly better than manual de-identification \cite{douglass2004computer}.  
However, these have been studied mostly in the single-institution setting. There are limited studies that evaluate de-identification performance of these methods across institutions \cite{yang2019study}. These suggest that NLP methods for de-identification perform poorly when evaluated on data from a different institution compared to the one that contributed the training data. This is especially significant in the context of patient data originating within Indian healthcare institutions. To the best of our knowledge, studies evaluating the performance of NLP based de-identification systems on patient data from Indian healthcare institutions have not yet been carried out. One reason for this might be that until recently there was no regulatory framework for accessing patient data for research. The Indian Digital Personal Data Protection Act 2023 (DPDPA) \cite{dpdpa2023} is a landmark legislation that came into effect in September 2023 and covers all organizations that process the personal data of individuals in India. Similar to GDPR 2016, the DPDPA defines responsibilities for organizations that collect, store, and process data from patients in India and holds them legally accountable for safeguarding patient privacy. The DPDPA also highlights the need for a data de-identification solution that has been validated on patient data from Indian healthcare institutions.

\noindent The present study takes a step towards answering this imminent need. Using a dataset of fully de-identified 99 discharge summaries obtained under Institutional Review Board (IRB) approval from the Sanjay Gandhi Post Graduate Institute of Medical Sciences (SGPGIMS), Lucknow, India, the study evaluates language models (LMs) for the task of de-identification. Furthermore, commercially available de-identification solutions are also evaluated. Hereafter, we refer to this dataset as the Indian Clinical Discharge Summaries ({\dataReal}, subscript $R$ refers to real) dataset. Given the paucity of clinical data, the study also evaluates Large Language Models (LLMs) on the task of generating synthetic clinical texts for training de-identification models. Critically, the study highlights the existence of several personal health information (PHI) elements in the {\dataReal}\   dataset that are unique to the language use and cultural practices in India. It is unlikely that the existing de-identification solutions have been trained to recognize these unique PHI elements, and therefore, their detection may be unreliable.
In a nutshell, we make the following contributions:
\begin{itemize}[noitemsep,nosep] 
\item We introduce a new dataset (Indian Clinical Discharge Summaries (\dataReal)) obtained from an Indian hospital and evaluate the performance of \model\ model \cite{huggingfaceXoocaroberta_ner_personal_info_10Hugging} (fine-tuned on non-Indian clinical summaries) on \dataReal\ for the task of De-Identification. Our experiments show poor cross-institutional performance. Experiments with existing commercial off-the-shelf clinical de-identification systems show similar trends. 
\item To overcome the paucity of Indian clinical data, we generate synthetic summaries using LLMs (Gemini \cite{team2023gemini}, Gemma \cite{team2024gemma}, Mistral \cite{jiang2023mistral}, and Llama3 \cite{touvron2023llama}) via In-Context Learning (ICL). Further, the synthetic summaries are used to train \model\ for de-identification on \dataReal. Results show significant improvement in the performance of the de-identification system. 
\item We release the model code and experiments via GitHub: \url{https://github.com/Exploration-Lab/llm-for-clinical-report-generation-deidentification}
\end{itemize}

\vspace{-2mm}
\section{Related Work} \label{sec:related}
\vspace{-2mm}

Automatic data de-identification methods for biomedical texts have focused on leveraging machine learning techniques to ensure privacy while maintaining data utility. Named Entity Recognition (NER) systems have been tailored to identify and anonymize personal health information/personal identifiable information (PHI/PII) from clinical narratives. Earlier work explored Support Vector Machines (SVMs) for identifying PHI \cite{neamatullah2008automated}. Researchers have also explored deep learning models, such as Convolutional Neural Networks (CNNs) and Recurrent Neural Networks (RNNs) \cite{dernoncourt2017identification}, which have shown superior performance over the conventional approach. 

\begin{figure*}[t]
  \centering
  \includegraphics[scale=0.5]{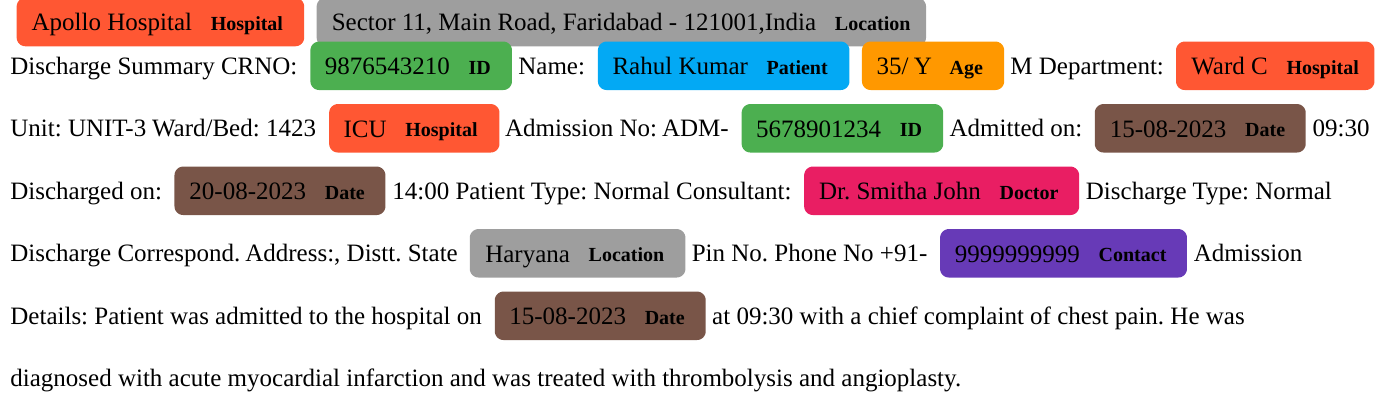}
  \caption{A sample of annotated text from Discharge Summary}
  \label{fig:Annotation_example}
  \vspace{-4mm}
\end{figure*}

\noindent In recent years, there has been a growing interest in the application of transformer-based models like BERT \cite{devlin2018bert} for the clinical NER and de-identification task \cite{chaudhry2022reducing,alsentzer2019publicly}. LLMs have also been explored for various clinical tasks such as clinical NLI \cite{mandal2024iitk}. Hybrid approaches that combine rule-based and machine learning methods have also been developed to enhance the robustness of de-identification systems \cite{meystre2010text}. A study by \citet{yang2019study} used a hybrid model combining Long Short-Term Memory (LSTM) networks with Conditional Random Fields (CRFs) for the de-identification of clinical notes. 
It demonstrated the effectiveness of integrating local resources and diverse word embeddings, and achieved high F1 scores across various de-identification tasks. Furthermore, \citet{elazzouzi2023automatic} de-identified French electronic health records using distant supervision and deep learning techniques. The study utilized models like Bi-LSTM+CRF and enhanced them with contextualized word embeddings. It achieved remarkable accuracy in removing identifiable information while maintaining data utility. These innovations underscore the continuous improvement and adaptation of de-identification methods to address the evolving challenges in data privacy. With remarkable progress made in generative AI techniques, researchers have started exploring generating synthetic clinical data. For example, medGAN \cite{choi2017generating} has been proposed to generate high-dimensional discrete variables such as patient records. It shows that it can produce realistic EHR data that preserves the statistical properties of the original dataset. Researchers have also explored differential privacy techniques in conjunction with  Generative Adversarial Networks (GANs) to ensure that the synthetic data does not allow re-identification of individuals. There is also ongoing research into hybrid models that combine rule-based and machine learning techniques to generate data that not only looks realistic but also adheres to known clinical correlations and constraints \cite{isasa2024comparative,goncalves2020generation}. Such approaches ensure that the synthetic data is both safe and scientifically valid for use in biomedical modeling simulations. The trend highlights the potential of synthetic data to address privacy and data availability challenges in biomedical research. In this paper, we explore LLMs for generating synthetic clinical reports that closely resemble reports in \dataReal, thus capturing the underlying data generation processes. 

\section{Clinical Discharge Summaries Datasets} \label{sec:real-corpus}

\noindent\textbf{n2c2:}
We make use of the 2006 and 2014 n2c2 datasets \cite{Uzuner2008,Stubbs2015}. The 2006 challenge involved the development of automated methods to de-identify discharge summaries from patient medical records \cite{Uzuner2008}. 
The total number of summaries in the \ncOld\ dataset are $888$, split between training and test sets. The 2014 challenge comprised of two tasks: de-identification and heart disease risk factor identification \cite{Stubbs2015}. For the de-identification task, the dataset included a variety of clinical documents such as progress notes, discharge summaries, and other narrative texts that typically contain detailed patient information. 

\noindent\textbf{Indian Clinical Discharge Summaries (\dataReal):}
We obtained fully de-identified 99 discharge summaries obtained under Institutional Review Board (IRB) approval from the Sanjay Gandhi Post Graduate Institute of Medical Sciences (SGPGIMS), Lucknow, India. All discharge summaries in the Indian Clinical Corpus were manually annotated for de-identified entities by human annotators using Doccanno \cite{doccano}, a data annotation tool. Each document was annotated by one annotator. The annotators had previous experience in clinical text annotations. Following established practice, we used the BIO scheme \cite{ramshaw1999text} for annotating named entities. Our PHI labels were defined by augmenting the PHI entities defined in the HIPAA Privacy Rule 2003 along with adaptation to Indian clinical texts. After annotation, we obtained 26 PHI unique entities in the {\dataReal} dataset. Subsequently, due to privacy concerns, PHI elements were replaced with fake values through an automatic replacement tool developed using the Python library \textit{Faker} \cite{Faraglia_Faker} (example in Fig. \ref{fig:Annotation_example}). Repeated occurrences of an entity within a note were tracked for consistent replacements. Moreover, settings such as date/time offsets were parameterized via a configurable file. The tool provides a scalable solution for de-identifying medical datasets while ensuring secure data access. Table \ref{tab:data-stats} provides statistics of the datasets. 

\begin{table*}[t]
\centering
\small
\renewcommand{\arraystretch}{1}
\setlength\tabcolsep{3pt}
\resizebox{\linewidth}{!}{
\begin{tabular}{@{}ccccccccc@{}}
\toprule
\multirow{2}{*}{Statistics}   & \multicolumn{5}{c}{Training dataset} & \multicolumn{3}{c}{Test set} \\ \cmidrule(lr){2-9}
 & \ncOld & \ncNew & \dataReal & \dataGenGemini & \dataGenLlama & \ncOld & \ncNew & \dataReal \\
 \cmidrule(lr){1-1} \cmidrule(lr){2-6} \cmidrule(lr){7-9}
\# Summaries & 668 & 790 & 79 & 1596 & 1043 & 220 & 514 & 20 \\
\# Unique Tokens & 29218 & 55907 & 13542 & 56780 & 25184 & 15231 & 41066 & 6106 \\
Max Length & 3023 & 2984 & 9494 & 4256 & 2590 & 2687 & 2474 & 8511 \\
Min Length & 13 & 74 & 97 & 100 & 109 & 15 & 99 & 270 \\
Avg. Summary Length & 581.71 & 618.86 & 1005.94 & 373.80 & 392.34 & 748.22 & 615.19 & 1343.40 \\
Original Tag Set & 9 & 24 & 26 & 34 & 106 & 9 & 21 & 24 \\
Mapped Tag set & 9 & 9 & 9 & 9 & 9 & 9 & 9 & 9  \\ \bottomrule
\end{tabular}
}
\caption{Statistics of various datasets}
\label{tab:data-stats}
\vspace{-4mm}
\end{table*}

\section{Generated Discharge Summaries Datasets} \label{sec:syngen}
Initial experimentation showed over-fitting in models on the \dataReal\ data due to its small size ($69, 10, 20$ summaries for train, val, and test sets, respectively). Consequently, we generated synthetic summaries to augment \dataReal\ data. Synthetic patient data is being used increasingly for a variety of in-silico biomedical experiments in addition to training data augmentation \cite{chen2021synthetic}.
\noindent Using the samples from \dataReal\, we generated medical discharge summaries specific to Indian patients using LLMs (Gemma, Llama-3-8B-Instruct, and Mistral-7B-Instruct-v0.1) via In-Context Learning (ICL). We experimented extensively with various prompts and discharge summaries, as explained below. Our choice of LLMs was driven by the feasibility of instantiating these models on-premise. Prompting is a key aspect of using LLMs. As described below, we experimented with various prompt designs.

\noindent \textbf{Discharge Summaries Generation using the \ncOld\ dataset:} Since the \ncOld\ discharge summaries are publicly accessible, we generated synthetic discharge summaries based on these along with PHI annotations using Gemini-pro-1.0. We arrived at a functional prompt by iteratively tuning and inspecting the synthesized outputs for overall length, presence of key subsections, and correct PHI annotation. While tuning our prompts, we did not check for the medical validity of the discharge summaries (see App. Table \ref{tab:prompts-for-ds-gen}). The prompt also contained an original \ncOld\ summary as an exemplar. This way, we generated five patient discharge summaries for each original discharge summary in the \ncOld\ dataset and a total of 3000 discharge summaries. The generated summaries were manually reviewed, and the ones containing gibberish text and missing or incorrect annotations were filtered out, resulting in 1596 synthetic discharge summaries with PHI annotations. Hereinafter, we refer to this dataset as \dataGenGemini\ .

\noindent \textbf{Discharge Summaries Generated using the \dataReal\ dataset:} The \dataReal\ dataset is accessible only under the Institutional Review Board’s approval, and therefore, LLMs that can be inferred only via public API endpoints cannot be used to process these. Consequently, we generated synthetic discharge summaries for the \dataReal\ dataset only with LLMs that could be instantiated within our secure compute infrastructure (Llama-3-8B-Instruct, Gemma and Mistral-7B-Instruct-v0.1, respectively). We evaluated various LLM and prompt combinations to converge on Llama-3-8B-Instruct (see App. Table \ref{tab:prompts-for-ds-gen} for the prompt). To evaluate the performance of model-prompt combinations, we calculated two metrics: BERT F1-Score and the average length of summaries (in words). The BERT F1-Score was calculated on a sample of synthetic annotated discharge summaries (target) and the 99 original \dataReal\ discharge summaries (see Table \ref{tab:model-prompt-comparison}). The BERT F1-Score of Meta-Llama-3-8B-Instruct and Mistral-7B-Instruct-v0.1 models with prompt B surpass other model-prompt combinations. We selected the Meta-Llama-3-8B-Instruct model for synthetic discharge summary generation and PHI annotation since, in addition to a high BERT F1-score, the generated summaries are, on average, longer. The \dataReal\ dataset was split so that $79$ summaries were used in the prompt to generate synthetic summaries while the remaining $20$ were reserved for the test set. The temperature parameter of Meta-Llama-3-8B-Instruct was set to $0.9$. Around $25$ summaries were generated for each of the $79$ \dataReal\ discharge summaries by embedding these one at a time as an exemplar in the prompt. In total, $1831$ discharge summaries, which already had PHI annotations, were generated, yielding $1043$ generated discharge summaries after manual review and filtration. Hereinafter, we refer to this dataset as \dataGenLlama. Further, we asked two annotators to annotate 50 generated summaries (after removing the PHI tags) with PHI tags. The Cohen's kappa coefficient \cite{cohens-kappa}, the measure of inter-annotator agreement, was 0.921, showing a high agreement. 

\noindent \textbf{Evaluation of the Quality of the Generated Summaries:} We assessed the face validity of the generated summaries by asking physicians to review a convenience sample of 30 real and 30 synthetic discharge summaries with the real/synthetic labels suppressed. The 60 discharge summaries were shuffled and uploaded to a secure, online review tool accessible only to the reviewers (physicians). The reviewers were asked to review each summary and then assign a single label (real or synthetic) to each based on their experience. The review results were compiled, and the precision, recall, and F1 scores were computed for each physician along with Cohen's Kappa to assess agreement between the two physicians (details in \S\ref{sec:exp}). 

\noindent As can be observed in Table \ref{tab:data-stats}, for the purpose of uniformity and modeling, we mapped PHI entities in each of the dataset to 9 tags (corresponding to 8 unique entities + 1 \texttt{OTHERS}). App. Table \ref{tab:TagMapping} provides details of tag mapping where the PHI entities are mapped with to their superset and all non-PHI entities are mapped to \texttt{OTHERS} Tag.

\begin{table}[t]
    \centering
    \small
    \renewcommand{\arraystretch}{1}
\setlength\tabcolsep{3pt}
\resizebox{\linewidth}{!}{
    \begin{tabular}{cccc}
        \hline
        \textbf{Prompt Id} & \textbf{Model Used} & \textbf{BERT F1-Score} & \textbf{Avg. Summary}  \\
        &   &  &   \textbf{Length (words)}\\
        \midrule
        B & meta-llama/Meta-Llama-3-8B-Instruct  & 0.491 & 564\\
        B & mistralai/Mistral-7B-Instruct-v0.1 & 0.493 &  400 \\
                C & meta-llama/Meta-Llama-3-8B-Instruct & 0.486 & 503\\
        C & mistralai/Mistral-7B-Instruct-v0.1 & 0.468 &  267\\
                C & google/gemma-1.1-7b-it & 0.478 & 268 \\
        \hline
  \end{tabular}
  }
    \caption{Comparison of model-prompt combinations }
    \label{tab:model-prompt-comparison}
\end{table}

\section{De-Identification Task} \label{sec:method}

\begin{table}[t]
\tiny
\centering
\begin{tabular}{@{}cccc@{}}
\toprule
\multirow{2}{*}{\textbf{Training Set}} & \multicolumn{3}{c}{\textbf{Test Set}} \\  \cmidrule(lr){2-4} 
  & \ncOld & \ncNew & \dataReal \\ \cmidrule(lr){1-1} \cmidrule(lr){2-4}
\ncOld & \checkmark & \checkmark & \checkmark \\
\ncNew & \checkmark & \checkmark & \checkmark \\
\ncOld + \ncNew & \checkmark & \checkmark & \checkmark \\
\dataGenGemini &  \checkmark & \checkmark & \checkmark \\
\dataGenLlama &  \checkmark & \checkmark & \checkmark \\
\dataGenGemini + \dataGenLlama & \checkmark & \checkmark & \checkmark \\
\dataGenGemini + \dataGenLlama + \ncNew & \checkmark & \checkmark & \checkmark \\
\dataGenGemini + \dataGenLlama + \ncNew & \checkmark & \checkmark & \checkmark \\ \bottomrule
\end{tabular}
\caption{Experiments Matrix}
\label{tab:experiments-matrix}
\vspace{-4mm}
\end{table}

\begin{table}[t] 
\centering
\tiny
\renewcommand{\arraystretch}{1}
\setlength\tabcolsep{3pt}
\begin{tabular}{@{}ccc@{}}
\toprule
\multirow{2}{*}{Attribute}  & \multicolumn{2}{c}{Dataset} \\ \cmidrule(l){2-3} 
 & Real & Generated \\ \cmidrule(l){1-1} \cmidrule(l){2-3}  
Counts & 3158684 & 5022667 \\
Length (words) & 560753 & 721886 \\
Mean $\pm$ SE & 4.64 $\pm$ 0.004  & 5.93 $\pm$ 0.005 \\
Median & 4.0 & 5.0 \\
Min & 1 & 1 \\
Max & 50 & 89 \\ \midrule
Jaccard Distance & \multicolumn{2}{c}{0.83} \\
BERTScore (F1) & \multicolumn{2}{c}{0.64} \\
BERTScore (Precision) & \multicolumn{2}{c}{0.65} \\
BERTScore (Recall) & \multicolumn{2}{c}{0.63} \\ \bottomrule
\end{tabular}
\caption{Comparison of {\ncOld} and {\dataGenGemini} Dataset}
\label{tab:datasetn2c2}
\end{table}

\begin{table}[t]
\centering
\tiny
\renewcommand{\arraystretch}{1}
\setlength\tabcolsep{3pt}
\begin{tabular}{@{}ccc@{}}
\toprule
Attribute & \multicolumn{2}{c}{Dataset} \\ \cmidrule(l){2-3} 
 & Real & Generated \\ \midrule
Counts & 636805 & 4789863 \\
Length (words) & 102604 & 508244 \\
Mean $\pm$ SE & 5.21 $\pm$ 0.01 & 7.77 $\pm$ 0.01 \\
Median & 4.0 & 5.0 \\
Min & 1 & 1 \\
Max & 72 & 472 \\ \midrule
Jaccard Distance & \multicolumn{2}{c}{0.80} \\
BERTScore (F1) & \multicolumn{2}{c}{0.58} \\
BERTScore (Precision) & \multicolumn{2}{c}{0.60} \\
BERTScore (Recall) & \multicolumn{2}{c}{0.56} \\ \bottomrule
\end{tabular}
\caption{Comparison of {\dataReal} and {\dataGenLlama} Dataset}
\label{tab:dataset_ICSD-R}
\vspace{-4mm}
\end{table}

\begin{table*}[t]
\centering
\small
\renewcommand{\arraystretch}{1}
\setlength\tabcolsep{3pt}
\resizebox{\linewidth}{!}{
\begin{tabular}{@{}cccccccccc@{}}
\toprule
Training Data & \multicolumn{3}{c}{ {\ncOld} } & \multicolumn{3}{c}{{\ncNew} } & \multicolumn{3}{c}{{\ncOld} + {\ncNew} } \\ \midrule
Testing Data & {\ncOld} & {\ncNew} & {\dataReal} & {\ncOld} & {\ncNew} & {\dataReal} & {\ncOld} & {\ncNew} & {\dataReal} \\ \cmidrule(lr){1-1} \cmidrule(lr){2-4} \cmidrule(lr){5-7} \cmidrule(lr){8-10}
\texttt{CONTACT} & 0.98 & 0.66 & 0.18 & 0.73 & 0.95 & 0.20 & 0.96 & 0.93 & 0.24 \\
\texttt{PATIENT} & 0.95 & 0.65 & 0.81 & 0.82 & 0.98 & 0.85 & 0.91 & 0.96 & 0.77\\
\texttt{DOCTOR} & 0.95 & 0.89 & 0.64 & 0.93 & 0.98 & 0.76  & 0.97 & 0.98 & 0.54 \\
\texttt{ID} & 0.99 & 0.55 & 0.64 & 0.96 & 0.97 & 0.65 & 1.00 & 0.96 & 0.93\\
\texttt{DATE} & 0.98 & 0.43 & 0.16 & 0.70 & 0.99 & 0.97 & 0.97 & 0,98 & 0.97\\
\texttt{LOCATION} & 0.89 & 0.80 & 0.71 & 0.78 & 0.95 & 0.80 & 0.81 & 0.94 & 0.75\\
\texttt{HOSPITAL} & 0.94 & 0.79 & 0.34 & 0.87 & 0.94 & 0.36 & 0.94 & 0.93 & 0.40\\
\texttt{AGE} & 0.80 & 0.00 & 0.00 & 0.02 & 0.99 & 0.48 & 0.12 & 0.94 & 0.53\\
\midrule
Micro Avg & 0.96 & 0.66 & 0.41 & 0.81 & 0.98 & 0.80 & 0.96 & 0.97 & 0.80 \\
Macro Avg & 0.93 & 0.60 & 0.43 & 0.72 & 0.97 & 0.63 & 0.83 & 0.95 & 0.64\\
Weighted Avg & 0.96 & 0.61 & 0.31 & 0.84 & 0.98 & 0.78 & 0.96 & 0.97 & 0.78\\ \bottomrule
\end{tabular}
}
\caption{F1 scores for PHI entities with overall micro Avg F1 , macro Avg F1 , Weighted Avg F1}
\label{tab:F1-Score-n2c2}
\vspace{-2mm}
\end{table*}

\begin{table*}[h!]
\centering
\small
\renewcommand{\arraystretch}{1}
\setlength\tabcolsep{3pt}
\resizebox{\linewidth}{!}{
\begin{tabular}{@{}cccccccccc@{}}
\toprule
Training Data & \multicolumn{3}{c}{\dataGenGemini} & \multicolumn{3}{c}{\dataGenLlama} & \multicolumn{3}{c}{\dataGenGemini + \dataGenLlama}\\ \midrule
Testing Data & {\ncOld} & {\ncNew} & {\dataReal} & {\ncOld} & {\ncNew} & {\dataReal} & {\ncOld} & {\ncNew} & {\dataReal} \\  \cmidrule(lr){1-1} \cmidrule(lr){2-4} \cmidrule(lr){5-7} \cmidrule(lr){8-10}
\texttt{CONTACT} & 0.80 & 0.47 &  0.11& 0.55 & 0.38 & 0.96 & 0.93 & 0.67 & 0.98\\
\texttt{PATIENT} & 0.74 & 0.56 &  0.68& 0.05 & 0.32 & 0.95 & 0.83 & 0.60 & 0.90\\
\texttt{DOCTOR} & 0.86 & 0.78 &  0.88& 0.35 & 0.71 & 0.98 & 0.86 & 0.76 & 0.98\\
\texttt{ID} & 0.87 & 0.58 &  0.51& 0.81 & 0.61 & 1.00 & 0.93 & 0.63 & 0.98\\
\texttt{DATE} & 0.87 & 0.90 &  0.88& 0.70 & 0.84 & 0.99 & 0.90 & 0.88 & 0.99 \\
\texttt{LOCATION} & 0.71 & 0.78 &  0.34& 0.50 & 0.66 & 0.97 & 0.75 & 0.81 & 0.96 \\
\texttt{HOSPITAL} & 0.87 & 0.72 &  0.31& 0.42 & 0.51 & 0.97 & 0.88 & 0.70 & 0.98\\
\texttt{AGE} & 0.02 & 0.67 &  0.51& 0.02 & 0.38 & 0.96 & 0.06 & 0.56 & 0.97\\
\midrule
Micro Avg & 0.85 & 0.77 &  0.68& 0.55 & 0.67 & 0.98 & 0.88 & 0.76 & 0.98\\
Macro Avg & 0.72 & 0.68 &  0.53& 0.42 & 0.55 & 0.97 & 0.77 & 0.70 & 0.97\\
Weighted Avg & 0.86 & 0.77 &  0.69& 0.52 & 0.66 & 0.98 & 0.88 & 0.77 & 0.98\\ \bottomrule
\end{tabular}
}
\caption{F1 scores for PHI entities for the \model\ trained on generated data.}
\label{tab:F1-Score-Generated-Data}
\vspace{-4mm}
\end{table*}

\begin{table*}[t]
\centering
\small
\begin{tabular}{@{}ccccccc@{}}
\toprule
Training Data & \multicolumn{3}{c}{\ncNew + \dataGenLlama + \dataGenGemini} & \multicolumn{3}{c}{\ncNew + \ncOld + \dataGenGemini + \dataGenLlama} \\ \midrule
Testing Data  & {\ncOld} & {\ncNew} & {\dataReal} & {\ncOld} & {\ncNew} & {\dataReal} \\ \cmidrule(lr){1-1} \cmidrule(lr){2-4} \cmidrule(lr){5-7}
\texttt{CONTACT} & 0.89& 0.95& 0.98&  0.97&  0.96&  0.98\\
\texttt{PATIENT}  & 0.87& 0.97& 0.88&  0.94&  0.96&  0.88\\
\texttt{DOCTOR}  & 0.95& 0.97& 0.98&  0.98&  0.98&  0.99\\
\texttt{ID} & 0.99& 0.97& 0.99&  0.99&  0.96&  0.99\\
\texttt{DATE}  & 0.82& 0.99& 0.99&  0.99&  0.99&  0.99\\
\texttt{LOCATION}  & 0.76& 0.95& 0.98&  0.85&  0.94&  0.98\\
\texttt{HOSPITAL} & 0.93& 0.94& 0.96&  0.96&  0.94&  0.97\\
\texttt{AGE} & 0.02& 0.97& 0.96&  0.35&  0.97&  0.86\\
\midrule
Micro Avg  & 0.88 & 0.97& 0.97&  0.97&  0.97&  0.97\\
Macro Avg  & 0.78& 0.96& 0.96&  0.88&  0.96&  0.96\\
Weighted Avg  & 0.90& 0.97& 0.97&  0.97&  0.97&  0.97\\ \bottomrule
\end{tabular}
\caption{F1 scores of PHI entities when {\model} is fine-tuned on combination of datasets}
\label{tab:CombinationofGenerated+n2c2-data}
\vspace{-2mm}
\end{table*}

\noindent\textbf{De-Identification Task:} De-Identification is conceptually similar to a Named Entity Recognition task. Both \dataGenGemini\ and \dataGenLlama\ were pre-processed and converted into BIO format as is customary in Named Entity Recognition development (also see App. Fig. \ref{fig:Annotation_example_BI}). Formally, given some text, $S = (w_1, w_2, w_3,..., w_n)$ containing $n$ words, de-identification requires labeling each of the word $w_i$ with a tag $t_k$ coming from a NER tagset $t_1, t_2,..., t_T$. Subsequently, the labeled entities can be redacted or replaced with fake values for privacy protection. 

\noindent\textbf{De-Identification Model:} 
We fine-tuned several different NER models, including ghadeermobasher/BCHEM4-Modified-BioBERT-v1 \cite{huggingfaceGhadeermobasherBCHEM4ModifiedBioBERTv1Hugging} and Clinical-AI-Apollo/Medical-NER \cite{huggingfaceClinicalAIApolloMedicalNERHugging}. In each case, we used a training partition of the data to train and a validation partition for evaluation. However, the Clinical NER models did not perform well since they are designed to label medical entities such as disease, drugs, procedures, and devices (see App. \ref{app:model-training}). RoBERTa-NER-Personal-Info model \cite{huggingfaceXoocaroberta_ner_personal_info_10Hugging} showed good performance on \ncOld\ and \ncNew\ datasets. The architecture for {\model} is shown in App. Fig. \ref{fig:model-architecture}. \model\ is a 24-layered transformer model that predicts a label for each token.

\section{Model Training Experiments} \label{sec:modelexp}
Initial experiments with \dataReal\ using a 69-10-20 (train-val-test) split resulted in overfitting given that \dataReal\ is small, containing only $99$ discharge summaries. We also experimented with training the model on \ncOld\ and \ncNew\ datasets and testing on \dataReal\ to check for cross-institutional generalization. We experimented with several combinations of real and synthetic datasets and evaluated on the test set of \ncOld, \ncNew, and \dataReal. Table \ref{tab:experiments-matrix} shows the experiments matrix, in total we evaluated 24 different combinations. For all the experiments, we reserved $20$ summaries of \dataReal\ for testing. Note that these summaries were also not used for generation. For each experiment, \model\ was fine-tuned on each training set as given in Table \ref{tab:experiments-matrix} and tested on each corresponding test set. Details about training are given in App. \ref{app:model-training}

\begin{table}[t]
    \centering
    \small
    {
    \begin{tabular}{@{}ccc@{}}
        \hline
         Metric & AWS & GCP \\
        \hline
        F1 Score & 0.37 & 0.47\\
        \hline
    \end{tabular}}
    \caption{Results: AWS vs. GCP Solutions on \dataReal\ test set}
    \label{tab:sgpgi_aws_gcp_comparision}
    \vspace{-4mm}
\end{table}

\begin{table}[t]
    \centering
    \small
    {
    \begin{tabular}{@{}ccc@{}}
        \hline
        Entity& AWS & GCP \\
        \hline
        \texttt{DATE} & 0.39 &0.56 \\
        \texttt{NAME} & 0.57 &0.52 \\
        \texttt{LOCATION} & 0.20 & 0.22 \\
        \texttt{AGE} & 0.12 & 0.00\\
        \texttt{ID} & 0.17 & 0.17\\
        \texttt{CONTACT} & 0.63 & 0.36 \\
        \hline
    \end{tabular}}
    \caption{F1 scores for Entity-Wise Comparison of AWS and GCP Solutions on \dataReal\ test set}
    \label{tab:sgpgi_aws_gcp_comparision_entity}
    \vspace{-5mm}
\end{table}

\noindent\textbf{Comparison with Commercial De-Identification Systems}: We compared the performance of these on the \dataReal\ test set. In particular, we evaluated AWS’s (Amazon Web Services) Comprehend Medical DetectPHI \cite{comprehend_medical_docs} and GCP’s (Google Cloud Platform) Data Loss Protection (DLP) \cite{googleclouddeidentify} de-identification solutions. For comparison and evaluation, \dataReal\ test set was mapped to a common tag set, which includes \texttt{DATE}, \texttt{NAME}, \texttt{LOCATION}, \texttt{AGE}, \texttt{ID}, and \texttt{CONTACT}. To ensure consistency across the dataset, pre-processing steps were applied. For instance, titles such as `Dr.' and `Mr.' were removed from \texttt{NAME} entities in the \dataReal\ test set due to the solution's inability to recognize them. Certain tags and entities were excluded from the analysis to align with a common tag set. The \texttt{LOCATION} entity was standardized by merging all location-related entities (street, city, state, zip) into a single \texttt{LOCATION} entity. Similarly, \texttt{HOSPITAL}, \texttt{ORGANISATION\_NAME} and \texttt{ADDRESS} entities were consistently mapped to \texttt{LOCATION}.

\noindent\textbf{De-identification using LLMs:} We further evaluated the performance of LLMs on \dataReal\ test set. Meta-Llama-3-8B-Instruct was instantiated within our secure compute infrastructure, and the prompt was developed for medical text de-identification using the iterative approaches described in the foregoing sections. 

\section{Experiments, Results and Analysis} \label{sec:exp}
\noindent\textbf{Comparison of datasets}: The total number of summaries in the \ncOld\ dataset are $888$, split between training and test sets, as shown in Table \ref{tab:data-stats}. 
The n-gram analysis of the \ncOld\ and \dataReal\ datasets reveals distinct linguistic patterns reflecting their unique clinical foci. The \ncOld\ dataset features unigrams like `patient,' `discharge,' and medication-related terms such as `mg' and `po' and bigrams like `mg po' and `discharge date,' highlighting a narrative centered on patient management and clinical processes as shown in App. Fig. \ref{fig:n2c2realngrams}. In contrast, the \dataReal\ dataset (as shown in App. Fig. \ref{fig:sgpgirealngrams}) shows a marked presence of terms such as `pm,' `days,' and `mgdl,' and bigrams and trigrams like `10 days,' `daily 10,' and `cr x ray,' suggesting an orientation towards experimental or lab-result oriented narratives, with a particular emphasis on procedural timelines and diagnostic procedures. Hence, \dataReal\ focuses on a broader scope involving diagnostics and treatment monitoring.


\noindent\textbf{Real versus Generated Datasets}

\noindent\textbf{{\dataGenGemini} vs {\ncOld}:} We analyzed the \ncOld\ and the synthetic \dataGenGemini\ discharge summaries in terms of summary statistics, Jaccard distance, and BERTScore (using the ``dmis-lab/biobert-v1.1'' model) as shown in Table \ref{tab:datasetn2c2} \cite{lee2020biobert, zhang2020bertscore}. The Jaccard distance suggests a high level of lexical dissimilarity between the datasets, indicating that the synthetic dataset introduces a significant degree of variation compared to the real dataset. While indicating some differences, an F1 score of $0.6362$ indicates the real and synthetic datasets have semantic overlap. An n-gram analysis of the top 10 unigrams, bigrams, and trigrams unveils the differences between the two datasets, yet also underscores their relevance to the task at hand as shown in App. Fig.\ref{fig:n2c2realngrams}, Fig.\ref{fig:n2c2synngrams}, Fig.\ref{fig:PHIn2c2ngrams}, and Fig.\ref{fig:PHIn2c2synngrams}. These metrics suggest that while the synthetic dataset is designed to be distinct enough to introduce useful variability, it retains a substantive semantic similarity to the real dataset. This balance is crucial when synthetic data is used for tasks such as model training, where the goal is to ensure that the model is not only trained on a diverse set of data but also remains relevant and effective when applied to real-world data. The high Jaccard distance combined with the moderate BERTScore indicates that the synthetic dataset achieves this objective by being similar enough to the real dataset to be useful, yet different enough to enhance the dataset's diversity and robustness.

\noindent\textbf{\dataGenLlama\ vs \dataReal:} Similar to the \ncOld\ and \dataGenGemini\ datasets, we analyzed the \dataReal\ and \dataGenLlama\ datasets with summary statistics, Jaccard distance, and BERTScore, as shown in Table \ref{tab:dataset_ICSD-R}. The Jaccard distance suggests lexical dissimilarity implying injection of new vocabulary in the generated discharge summaries. The n-gram analysis of the top 10 unigrams, bigrams, and trigrams shows these differences (App. Fig.\ref{fig:sgpgirealngrams}, Fig.\ref{fig:sgpgisynngrams}, Fig.\ref{fig:PHIsgpgirealngrams}, and Fig.\ref{fig:PHIsgpgisynngrams}). The BERTScore results indicate a moderate level of semantic similarity between the real and generated datasets. The metrics suggest that the generated dataset has greater lexical variety and incorporates some additional semantic constructs.

\begin{figure}[h]
  \centering
  \includegraphics[scale=0.25]{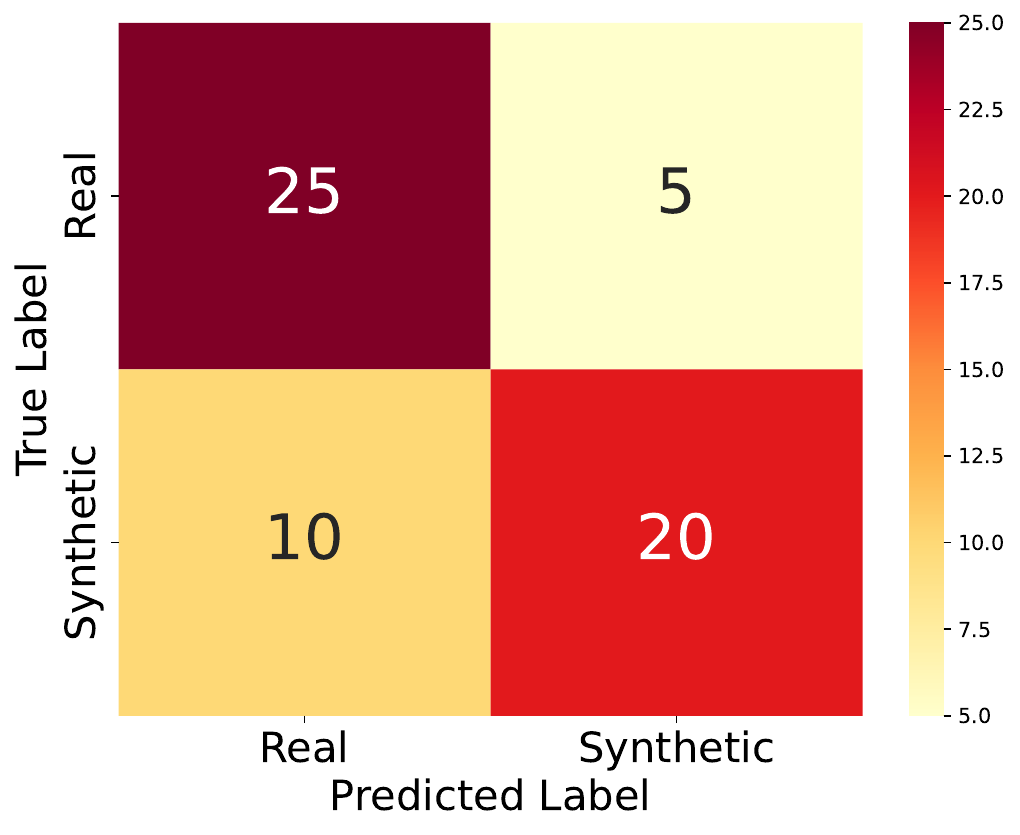}
  \caption{Confusion matrix on convenience sample (60 discharge summaries) evaluated by physician 1}
  \label{fig:cm_physician1}
\end{figure}

\begin{figure}[h]
  \centering
  \includegraphics[scale=0.25]{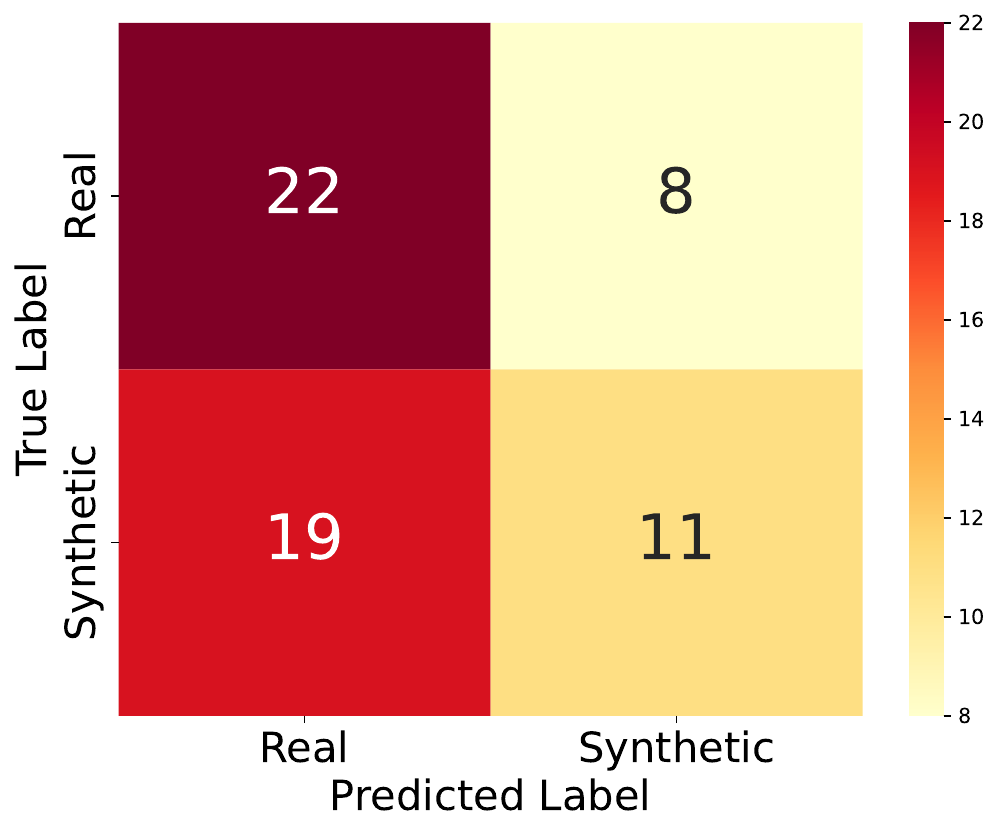}
  \caption{Confusion matrix on convenience sample (60 discharge summaries) evaluated by physician 2}
  \label{fig:cm_physician2}
  \vspace{-4mm}
\end{figure}

\noindent\textbf{Evaluation of The Quality of Generated Summaries}: The confusion matrix on convenience sample of 60 discharge summaries evaluated by physician1 and physician2 are shown in Fig. \ref{fig:cm_physician1} and Fig. \ref{fig:cm_physician2} respectively. There are 10 summaries that were originally synthetic but were labeled as real by physician 1, and 19 summaries that were originally synthetic but were labeled as real by physician 2. Physician 1 is able to label summaries with higher precision and recall, i.e., higher f1-score as compared to physician 2 (Table \ref{tab:em_physician1_physician2}). The Cohen's kappa coefficient, the measure of inter-annotator agreement, is $0.290$ showing a fair agreement between the labels assigned by the physicians. Additionally, physician 1 observed that many of the discharge summaries that he labeled synthetic appeared to have been translated from a non-English source. Physician 2 reported some diagnosis and formatting issues among the summaries he labeled as synthetic. Additionally, physician 2 reported some errors in diagnoses, medications, and lab results, but these were not limited to the summaries he labeled as synthetic.

\begin{table}[t]
    \centering
    \small
    \scalebox{0.9}{
    \begin{tabular}{llll}
        \hline
         Physician & Precision & Recall & F1-score \\
        \hline
        Physician 1 & 0.714 & 0.833 & 0.769   \\
        Physician 2 & 0.537 &  0.733 & 0.620  \\
        \hline
    \end{tabular}}
    \caption{Evaluation metrics of 60 discharge summaries annotated by physician 1 and physician 2}
    \label{tab:em_physician1_physician2}
    \vspace{-4mm}
\end{table}

\noindent\textbf{Model Performance:} Table \ref{tab:F1-Score-n2c2} shows the results for intra- and inter-institutional performance. As can be observed, the inter-institutional performance of the model is very high ($> 0.96$ F1). However, the cross-institutional performance suffers significantly. Table \ref{tab:F1-Score-Generated-Data} shows the results of training on generated datasets. The fine-tuned Model gives $68\%$ F1 score on the {\dataReal} test set, $77\%$ on the {\ncNew} test set, and $85\%$ on the {\ncOld}. Results on the {\dataReal} test set are not promising. This might have happened because {\dataGenGemini} was generated using {\ncOld}. Fine-tuning on {\dataGenLlama} dataset results in $98\%$ F1 score on the {\dataReal} test set, $67\%$ on {\ncNew} test set, and $55\%$ on the {\ncOld} test set. To further improve model generalization, we experimented with combinations of datasets. Table  \ref{tab:CombinationofGenerated+n2c2-data} shows the training results on a combination of real and synthetic datasets. We get micro-F1 of $97\%$ on {\ncNew} and {\dataReal} test set given that we have included {\ncNew} and {\dataGenLlama} datasets in training, but the performance of the model ($88\%$) is also notable on {\ncOld}  dataset. These results indicate that fine-tuning on the combination improves cross-institutional performance. 

\noindent\textbf{Analysis:} Our experiments indicate models have poor cross-institutional generalization. We performed several experiments  with  n2c2-2006, n2c2-2014, {\dataGenGemini}, and {\dataGenLlama} datasets, and their combinations. The general trend is that fine-tuned model performance degrades heavily in cross-dataset settings. At the individual entity level, the F1 score for the \texttt{PATIENT} entity is consistent for all the fine-tuned models. For the \texttt{DOCTOR} and \texttt{DATE} entities, the F1 scores of all the fine-tuned models are also consistent, except for when the model is trained on the n2c2-2006 dataset and tested on the n2c2-2014 dataset and {\dataReal} test sets. For the \texttt{ID} entity, all the fine-tuned models have consistent F1 scores, except for when the model is trained on {\dataGenGemini}, and tested on n2c2-2014 and {\dataReal} datasets. We noticed performance variance in the \texttt{LOCATION}, \texttt{AGE}, and \texttt{CONTACT} entities. This could be because the \texttt{LOCATION} can be any local address without a specific format.  \texttt{AGE} is either a number like `78 Y' or a word representation of that number like `Seventy-Eight year old'. In most cases in the datasets, these types of words or tokens are tagged as \texttt{OTHERS}, and they are highly prevalent. This could be why the \texttt{AGE} tag was incorrectly predicted as \texttt{OTHERS} in cross-dataset settings. The entity \texttt{CONTACT} includes email, IP address, phone number, landline number, etc. However, their distribution is not uniform. 

\noindent Our main aim was to develop a robust model that could de-identify medical text from Indian Healthcare Institutes. This was done by fine-tuning {\model} on {\dataGenLlama} where we are getting state-of-the-art performance on {\dataReal}. Almost all the entities were correctly identified, with a few exceptions. A few PHI entities were misidentified with non-PHI entities (i.e., \texttt{OTHERS} ) and vice versa, as can be seen in App. Fig. \ref{fig:cm_ICSD-G-l}.   
However, the percentage of incorrect prediction is significantly less when considering the total support set of {\dataReal} test set. However, this fine-tuned model was not generalizable when we tested it on the {\ncOld} and {\ncNew}  test sets, as seen in the Table \ref{tab:F1-Score-Generated-Data}. For model generalizability, we fine-tuned {\model} on {\ncNew} , {\dataGenGemini} and {\dataGenLlama}, tested on the {\ncOld} test set. The results shown in Table \ref{tab:CombinationofGenerated+n2c2-data} indicate that models are generalizing when we fine-tuned them on different combinations of datasets, although the F1 score for all entities is not consistent, as can be seen in App. Fig. \ref{fig:cm_n2c2_2014+ICSD-G-g+ICSD-G-l}. Confusion matrix for all the experiments are shown in App. Fig. \ref{fig:cm_2006} Fig. \ref{fig:cm_2014} , Fig. \ref{fig:cm_2006+2014} , Fig. \ref{fig:cm_ICDS-G-g} , Fig. \ref{fig:cm_ICSD-G-l} Fig. \ref{fig:cm_ICSD-G-l+ICSD-G-g} , Fig. \ref{fig:cm_n2c2_2014+n2c2_2006+ICSD-G-g+ICSD-G-l}. 

\noindent\textbf{Comparison with Commercial De-Identification Systems:} The results obtained using AWS and GCP solutions are summarized in Table \ref{tab:sgpgi_aws_gcp_comparision} and Table \ref{tab:sgpgi_aws_gcp_comparision_entity}. The results clearly indicate that AWS and GCP do not perform well on {\dataReal} test set. This could be because systems have been trained on non-Indian specific clinical data. This underscores the importance of ensuring that de-identification caters to diverse demographics, which is essential for ensuring the efficacy and ethical deployment of these solutions. 

\noindent The underperformance of commercial solutions in classifying PHI in \dataReal\ can be attributed to  misidentification. Medical entities are mistaken as  \texttt{NAME}/ \texttt{LOCATION}, while Pin-codes as  \texttt{ID}. Names like `Alia' and `Adah' are not being consistently recognized as  \texttt{NAME} by AWS and GCP. Patient IDs that start with CRNO: \#\#\#\#\#\#\#\#\#\#\# or ADM-\#\#\#\#\#\#\#\#\#\# are not identified as PHI; these solutions probably aren't sure what CRNO, ADM stand for. `B/O Kanav Viswanathan' is misidentified, where `Kanav Viswanathan' is a name and B/O stands for Baby of but gets labeled as a  \texttt{LOCATION}. `Urvi Bhamini Faiyaz Kakar' is identified as  \texttt{Name} by GCP but not by AWS. `Wockhardt Hospitals,' hospital name was not identified as PHI. Medical terms like `BILIRUBIN,' `MALLOY EVELYN,' `CR X Ray' and `SERUM LIPASE' are misidentified as  \texttt{NAME} when they describe medical tests. Similarly, `CREATININE (M - JAFFE COMPENSATED)' is a medical test and `JAFFE' is misidentified as  \texttt{NAME}. `Meropenem,' an antibiotic, is misidentified as  \texttt{NAME}. Even terms like `Ward' from room names such as `Ward-B' occasionally get misidentified as  \texttt{NAME}. Test results like `136/94mmHg' or `TSH - 5.45' are misidentified as  \texttt{ID}. Locations like `Subramaniam Chowk' and `Yohannan Nagar,' are also misidentified as  \texttt{NAME}. Additionally, using GCP or AWS for PHI detection introduces variability, causing results to vary with each execution. These factors underscore the need for precision and consistency in data handling to mitigate performance issues in medical contexts.





\noindent\textbf{De-identification using LLMs:} We also conducted experiments of de-identifying clinical summaries using LLMs directly. A precision score of $0.55$ was obtained. However, the model faced challenges in terms of recall. The recall scores were merely $0.11$. We also evaluated the performance of Mistral-7B-Instruct-346v0.1 and Gemma. Surprisingly, the results obtained from these models were far inferior to those of Meta-Llama-3-8B-Instruct. Results suggest that the LLMs struggle to detect PHI in Indian medical discharge summaries.

\section{Conclusion and Future Directions} \label{sec:conlusion}

In this paper, we explored the task of de-identification on Indian clinical discharge summaries. Experiments indicate a poor generalization of fine-tuned (on public datasets) models and poor performance of the off-shelf commercial systems. Experiments with LLM generated summaries look promising; the model fine-tuned on generated summaries and public datasets shows good generalization performance. 
Our results are based on a small test set.  
Using the insights from our work, we aim to set-up an active learning workflow that combines our fine-tuned model and human annotators to produce a larger test dataset on which we may evaluate overall model performance as well as by conditioning on a medical specialty. 
The augmented (generated summaries with original data) institution-specific dataset can be used to fine-tune NER models that have been pre-trained on PHI data cost-effectively. Achieving cross-institution portability remains a topic of active research. However, many open-source large language models can be deployed on-premise and, as described above, fine-tuned to provide an immediate and effective solution to personal data protection in Indian healthcare institutions. 

\section{Acknowledgements} 
We would like to thank Dr. Uttam Singh, Dr Prabhakar Mishra, and Dr Amit Goel for their support for this work.

\bibliography{references}

\clearpage
\newpage

\section*{Appendix} \label{sec:appendix}

\appendix


\titlecontents{section}[18pt]{\vspace{0.05em}}{\contentslabel{1.5em}}{}
{\titlerule*[0.5pc]{.}\contentspage} 

\titlecontents{table}[0pt]{\vspace{0.05em}}{\contentslabel{1em}}{}
{\titlerule*[0.5pc]{.}\contentspage} 

\startcontents[appendix] 
\section*{Table of Contents} 
\printcontents[appendix]{section}{0}{\setcounter{tocdepth}{4}} 

\startlist[appendix]{lot} 
\section*{List of Tables} 
\printlist[appendix]{lot}{}{\setcounter{tocdepth}{1}} 

\startlist[appendix]{lof} 
\section*{List of Figures} 
\printlist[appendix]{lof}{}{\setcounter{tocdepth}{1}} 

\newpage

\section{Prompts and Synthetic Discharge Summaries}
In Table \ref{tab:prompts-for-ds-gen}, we showcase the prompts which we used to generate the {\dataGenGemini} and {\dataGenLlama} datasets. We used Prompt A in Table \ref{tab:prompts-for-ds-gen} for generating {\dataGenGemini} from Gemini-pro-1.0. Table \ref{tab:example_generated_summaries_gemini} gives a sample discharge summary. Using prompt B in Table \ref{tab:prompts-for-ds-gen}, we generated {\dataGenLlama} dataset using Llama-3-8B-Instruct. Table \ref{tab:example_generated_summaries_llama} gives a sample discharge summary.
\begin{table*}[t]
    \centering
    \scalebox{0.76}{
    \begin{tabular}{c p{18cm}}
        \hline
        \textbf{Prompt Id} & \textbf{Prompt} \\
        \hline
        A &  Generate discharge summaries for Indian patients, capturing the essence of healthcare in India. The summaries should integrate conventional medical treatments with traditional remedies, reflecting the holistic approach embraced by Indian healthcare systems. Incorporate prevalent Indian health conditions, treatments, and culturally relevant follow-up care instructions. To ensure authenticity, each summary should include distinct patient details like name, age, address, contact information, hospital, doctor, and ID. Include prevalent diseases in India such as Tuberculosis (TB), Diabetes, Cardiovascular Diseases, Respiratory Infections, Hypertension, Dengue Fever, Malaria, Hepatitis, Chronic Kidney Disease (CKD), Cancer, Typhoid Fever, Cholera, HIV/AIDS, Japanese Encephalitis, Leptospirosis, Rabies, Tuberculosis of the Central Nervous System (CNS TB), Rheumatic Heart Disease, Iron-Deficiency Anemia, and Chikungunya. Also, laboratory test reports of the chosen disease should be included. Ensure the format of generated discharge summaries is similar to the summary given in the prompt, i.e., in XML format. 
        
        Example: Patient Summary: <discharge summary>  
        
        Generate the summaries that have a minimum of 2048 words. Ensure there is consistent consistency between the doctor's name, patient name, drug-disease, etc. \\ \\
       \hline 
        B & Generate an extensive discharge summary of at least 2048 words tailored for Indian patients. To ensure authenticity, the generated summary must include distinct patient-specific details like name, age, address, contact information, hospital name, doctor name, and unique ID. Maintain coherence across all the elements, doctor's name, patient’s identity, medications, diseases, etc. Ensure all the PHI (personal health information) elements are properly annotated to maintain privacy and authenticity.
        
        The generated discharge summary should be XML-formatted with PHI annotations. The generated summaries should include following sections:\ Admission Details, Diagnosis / Chief Complaints, Allergies, Physical Examination, Medical History, Family Medical history, Treatment Plan, Investigations, Medications (List of medications prescribed at discharge), Follow-up Instructions, Procedures/Lab Tests Conducted (List of procedures or tests conducted during hospital stay, along with results if available), and Special Instructions.
        
        Please ensure that these sections are incorporated into the generated summaries, but refrain from including them as tags in the output. The generated summary should be properly enclosed within the <RECORD> and </RECORD> tags to ensure it's within the XML format.
        
        Here's an example patient summary: 
        
        Patient Summary: <discharge summary>\\ \\
       \hline
       C & Generate an extensive synthetic discharge summary of at least 2048 words tailored for Indian patients. Generated summary must include distinct entities like name, age, address, contact information, hospital name, doctor name, and unique ID. Maintain coherence across all the elements, doctor's name, patient’s identity, medications, diseases, etc. Identify all entities in the generated text and mark these with XML tags as in the following example:<TYPE='Insurance Number'>AB123456C</TYPE>

        entities= [`Patient Name',`Hospital\_Name',`Staff\_Name',`Doctor\_Name',`Age',`Gaurdian\_Name',`Gender',
        
    `Patient\_ID',`Misc\_Medical\_ID',`Aadhar',`Driver\_License',`Voter\_ID',`PAN\_Card',`Patient\_DOB',
        
        `Treatment\_Date',`Treatment\_Time',`Phone\_No',`Landline',`Email',`IP\_Address',`Fax',`Doctor\_Specialisation',

        `Patient\_Profession',`City',`Ward\_Location',`Device\_Number',`Other\_Info',`State',`Street',`Zip',`Country',
        
        `Other\_Location',`Other\_Govt\_ID',`Insurance\_Number',`Web\_url']
        
        The generated summaries should include following sections: Admission Details, Diagnosis / Chief Complaints, Allergies, Physical Examination, Medical History, Family Medical history, Treatment Plan, Investigations, Medications (List of medications prescribed at discharge), Follow-up Instructions, Procedures/Lab Tests Conducted (List of procedures or tests conducted during hospital stay, along with results if available), and Special Instructions. Please ensure that these sections are incorporated into the generated summaries, but refrain from including them as tags in the output.
        
        The generated summary should be properly enclosed within the <RECORD> and </RECORD> tags to ensure it's within XML format.
        
        Here's an example synthetic patient summary:
        
        <discharge summary>. \\ \\
        \hline
    \end{tabular}}
    \caption{Prompts used for synthetic discharge summary generation}
    \label{tab:prompts-for-ds-gen}
    \end{table*}

\begin{table*}[t]
    \centering
    \resizebox{0.9\textwidth}{!}{
    \begin{tabular}{c p{16cm}}
        \hline
        \textbf{Model Name} & \textbf{Generated Discharge Summary} \\
        \hline
        Gemini-pro-1.0 &  829369
HODYEGE DISTRICT HOSPITAL
9151843
830690
408542
12/12/10/2023 23:59:00 PM
Discharge Summary
Discharge
Report Status :
Signed
DISCHARGE SUMMARY NAME :
NIVEDITHA CHETTY
UNIT NUMBER :
393-29-55
ADMISSION DATE :
12/10/10/2023
DISCHARGE DATE :
12/12/10/2023
PRINCIPAL DIAGNOSIS :
Prolapsed intervertebral disc with radiculopathy of L4-L5, L5-S1, L2-L3, with weakness of right lower limb.
ASSOCIATED DIAGNOSIS :
CVA (LEFT SIDED HEMIPARESIS)
ADVERSE DRUG REACTIONS :
None .
CLINICAL INFECTIONS :
None .
PRINCIPAL PROCEDURE :
Spinal surgery
ASSOCIATED PROCEDURES :
Physiotherapy .
HISTORY OF PRESENT ILLNESS :
Mrs. NIVEDITHA CHETTY is a 58 years old female who presented to the emergency department with complaints of low back pain for 1 year and weakness of right lower limb for 1 month . Weakness was insidious in onset and gradually increased to the present level . There was no numbness of lower limbs . No urinary or bowel involvement . No fever , cough , gastrointestinal symptoms or cardio-respiratory symptoms .
PAST MEDICAL HISTORY :
CVA (LEFT SIDED HEMIPARESIS) .
SOCIAL HISTORY :
Non smoker and non alcoholic .
PHYSICAL EXAMINATION :
VITALS :
Temperature : 98.6\textbackslash302\textbackslash260 F, BP : 120/80 mmHg , SPO2 : 95\% , Respiratory rate : 20/min , Pulse rate : 80/min
CNS :
GCS 15/15 , Cranial nerves - Normal . Power - Right lower extremity grade - 4/5 . Sensation - Intact . Motor system examination - Tone - Increased on the right side . Deep tendon reflexes - Brisk on the right side . Plantar - Extensor on the right side .
CVS :
No pallor , icterus , clubbing , pedal edema , cyanosis , SPO2 - 99\% . JVP not raised . Carotid pulsations are well felt and equal on both the sides . No bruits audible . Heart sounds - Regular . No murmurs , gallops , rubs or extra sounds .
Respiratory system :
Chest expansion - Symmetrical . Air entry reduced bilaterally but more on the right side . No added sounds . Breath sounds decreased in right infrascapular area .
Abdomen :
Soft , non-tender , no mass or organomegaly .
LABORATORY DATA :
Hb - 12.0 gm/dl , TLC - 8,000/cmm , Platelets - 2.5 lacs/cmm , ESR - 20 mm/hr . Blood sugar - 100 mg/dl , Urea - 30 mg/dl , Creatinine - 1.0 mg/dl , Uric acid - 5.0 mg/dl . Liver function tests - Normal . Lipid profile - Normal . Electrolytes - Normal .
X-RAY :
LUMBAR SPINE AP  LATERAL VIEW :
There is narrowing of the L4-L5, L5-S1, L2-L3 intervertebral spaces with sclerosis of the adjacent end plates . Loss of lumbar lordosis with mild dextroscoliosis .
CT SCAN :
LUMBAR SPINE :
There is a prolapsed intervertebral disc at L4-L5, L5-S1, L2-L3 with foraminal stenosis on the right side at L4-L5 . Thecal sac is compressed at L4-L5, L5-S1 due to central disc prolapse .
MRI SCAN :
LUMBAR SPINE :
There is a prolapsed intervertebral disc at L4-L5, L5-S1, L2-L3 with foraminal stenosis on the right side at L4-L5 . Thecal sac is compressed at L4-L5, L5-S1 due to central disc prolapse . There is evidence of denervation of the right sciatic nerve and cauda equina .
HOSPITAL COURSE :
The patient was admitted for surgical intervention . After proper pre-operative evaluation the patient was taken up for surgery . During surgery , a decompression of the thecal sac and foramina was done .
Postoperatively , the patient was started on physiotherapy . The patient was able to walk with the support of walker .
MEDICATIONS ON DISCHARGE :
Tramadol - 50 mg - BD SOS - 7 days
Aceclofenac - 100 mg - BD - 7 days
Multivitamin and minerals - OD
FOLLOW UP :
The patient will be reviewed in the clinic in 2 weeks .
DR.GYANESHWARI JANGID
MBBS, MD, MS (Orthopedics)
HODYEGE DISTRICT HOSPITAL
Electronically Signed
DR.GYANESHWARI JANGID
[ report\_end ]\\
\hline
\end{tabular}
}
    \caption{Example summary generated using gemini-pro-1.0}
    \label{tab:example_generated_summaries_gemini}
    \end{table*}

\begin{table*}[t]
    \centering
    \scalebox{0.76}{
    \begin{tabular}{c p{15cm}}
        \hline
        \textbf{Model Name} & \textbf{Generated Discharge Summary} \\
        \hline
        Meta-Llama-3-8B-Instruct &  Max Super Specialty Hospital
Sector 11, Dwarka, New Delhi - 110075,India Discharge Summary
    CRNO: 1234567890 Name: Rahul Kumar 35/Y/M Department: Ward E
    Unit: UNIT-3 Ward/Bed: 5432 ICU
    Admission No: ADM-9012345678 Admitted on: 25-08-2023 14:30 Discharged on: 28-08-2023 10:00
    Patient Type: Normal Consultant: Dr. Rohan Sharma Discharge Type: Normal Discharge
    Correspond. Address:, Distt. State Uttar Pradesh Pin No. Phone No +91-9812345678
    FOLLOW UP: Appointment on 15/10/2023 Morning Block-Upto 09:00 A.M.INVESTIGATION RESULTS: 
    25/08/2023 09:30 AM Blood Urea 40 mg/dL
    25/08/2023 09:30 AM Serum Creatinine 1.2 mg/dL
    25/08/2023 09:30 AM Hb 12.5 g/dL

    Admission Details:
    Patient was admitted to the hospital with a diagnosis of Acute Kidney Injury (AKI) due to dehydration. He was a known hypertensive and diabetic patient. He presented with symptoms of fatigue, weakness, and decreased urine output.

    Diagnosis/Chief Complaints:
    Primary Diagnosis: Acute Kidney Injury (AKI)
    Secondary Diagnosis: Hypertension, Diabetes Mellitus

    Allergies:
    None reported

    Physical Examination:
    The patient was conscious and oriented. He had a pulse rate of 100 beats per minute, blood pressure of 160/100 mmHg, and respiratory rate of 20 breaths per minute. He had bilateral pedal edema and decreased urine output.

    Medical History:
    The patient had a history of hypertension and diabetes mellitus. He was taking medications for these conditions, including metformin and lisinopril.

    Family Medical History:
    The patient's father had a history of hypertension and heart disease.

    Treatment Plan:
    The patient was treated with intravenous fluids and medications to manage his symptoms. He was also started on dialysis to manage his acute kidney injury.

    Investigations:
    The patient underwent the following investigations:
    Blood Urea 40 mg/dL
    Serum Creatinine 1.2 mg/dL
    Hb 12.5 g/dL

    Medications:
    The patient was prescribed the following medications at discharge:
    Metformin 500 mg twice daily
    Lisinopril 10 mg daily
    Furosemide 20 mg daily
    Hydrochlorothiazide 25 mg daily
    Aspirin 75 mg daily

    Follow-up Instructions:
    The patient was advised to follow up with his primary care physician in two weeks for a check-up and to monitor his blood sugar and blood pressure levels.

    Procedures/Lab Tests Conducted:
    The patient underwent the following procedures and lab tests during his hospital stay:
    Blood tests: Blood Urea 40 mg/dL, Serum Creatinine 1.2 mg/dL, Hb 12.5 g/dL
    Urine tests: Urine output decreased, proteinuria
    Dialysis: Three sessions of dialysis were performed

    Special Instructions:
    The patient was advised to follow a strict fluid intake and to monitor his urine output closely. He was also advised to take his medications as prescribed and to follow up with his primary care physician regularly.

    Prepared by (Dr. Rakesh Kumar ) Signature of Consultant Printed on 29-08-2023 14:30:15 Temp  Employee Male @ 192.168.1.100 Page 2 / 2    \\

                \hline
\end{tabular}}
    \caption{Example summary generated using llama-3-8B-Instruct}
    \label{tab:example_generated_summaries_llama}
    \end{table*}

\begin{figure*}[t]
  \centering
  \includegraphics[scale=0.3]{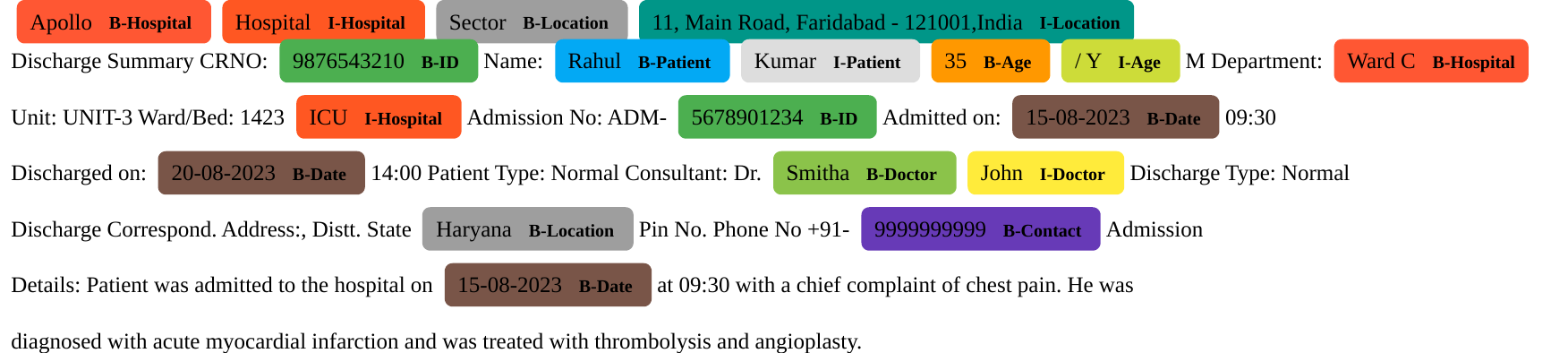}
  \caption{Pre-processed Discharge Summary after adding B and I tags}
  \label{fig:Annotation_example_BI}
\end{figure*}

\begin{table*}[t]
\centering
\small
\begin{tabular}{@{}lc@{}}
\toprule
\multicolumn{1}{c}{Original tags} & Mapped Tags \\ \midrule

\begin{tabular}[l]{@{}l@{}}Treatment\_Date, Patient\_DOB, Investigation\_Date, Admission Date, \\ Procedure\_Date, Date\end{tabular}  & \texttt{DATE}        \\
Ward\_Location, Hospital\_Name, Department  & \texttt{HOSPITAL}    \\
Patient\_ID, Misc\_Medical\_ID, Employee\_ID, Admission Number   & \texttt{ID}          \\
Age & \texttt{AGE}         \\
\begin{tabular}[l]{@{}l@{}}Doctor\_Name, Staff\_Name, Prepared by, Signature, Doctor\_Signature, \\ Signature of Consultant\end{tabular}              & \texttt{DOCTOR}      \\
\begin{tabular}[l]{@{}l@{}}Patient\_Name, Gaurdian\_Name, Patient\_Signature, Patient\_Spouse, \\ Family\_Member\_Name\end{tabular}      & \texttt{PATIENT}     \\
\begin{tabular}[c]{@{}l@{}}Zip, Phone\_No, Landline, IP\_Address, Phone, Contact\_Info,\\ Contact\_Number, Contact\_No, Mobile, Phone Number, Patient\_Phone, \\ Email, Email\_ID, Contact Information, Phone No\end{tabular} & \texttt{CONTACT}     \\
\begin{tabular}[c]{@{}l@{}}City, State, Country, Street, Other\_Location, Correspondence\_Address, \\ Contact\_Address, Contact Information, Pin, Pin Code, Pin\_No, \\ Postal\_Code,  Address, Contact\_Address\end{tabular}    & \texttt{LOCATION}   \\ \bottomrule
\end{tabular}%
\caption{Tag mapping from PHI entities in the different datasets to the PHI  entity set of {\ncOld} dataset, and \\ all other non-PHI entities are mapped with  \texttt{Others} tag}
\label{tab:TagMapping}
\end{table*}

\section{Tag Mapping across all the dataset and tag Distribution after Mapping the Tags }

We have five datasets: {\ncOld}, {\ncNew}, {\dataReal}, {\dataGenGemini}, and {\dataGenLlama}. Each dataset has its own tag set. {\ncOld} contains 9 tags, {\ncNew} contains 24, {\dataReal} contains 26, {\dataGenGemini} contains 34, and {\dataGenLlama} contains 106 unique tags, including the \texttt{OTHERS} tag. In the datasets {\ncOld}, {\ncNew}, and {\dataReal}, all the tags are related to PHI entities. However, in the {\dataGenLlama} and {\dataGenGemini} datasets, a few annotated tags are not related to the PHI entities due to LLM hallucinations. 
To train models for a fair comparison, we need a uniform tag set across all datasets.

\begin{figure}[t]
  \centering
  \includegraphics[scale=0.18]{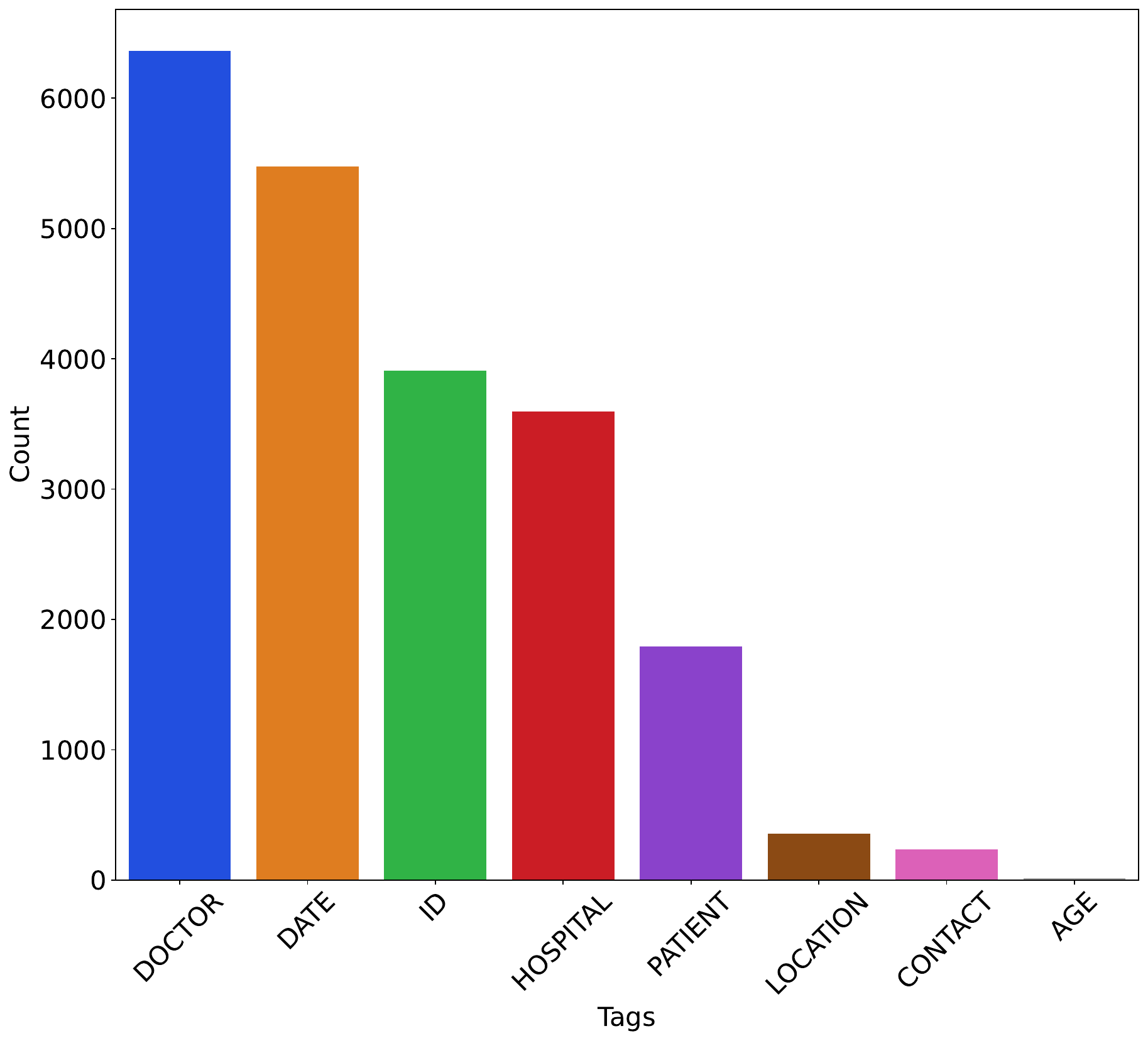}
  \caption{Tag Distribution in {\ncOld} train dataset }
  \label{fig: Tag-n2c2-2006-train}
\end{figure}

\begin{figure}[t]
  \centering
  \includegraphics[scale=0.18]{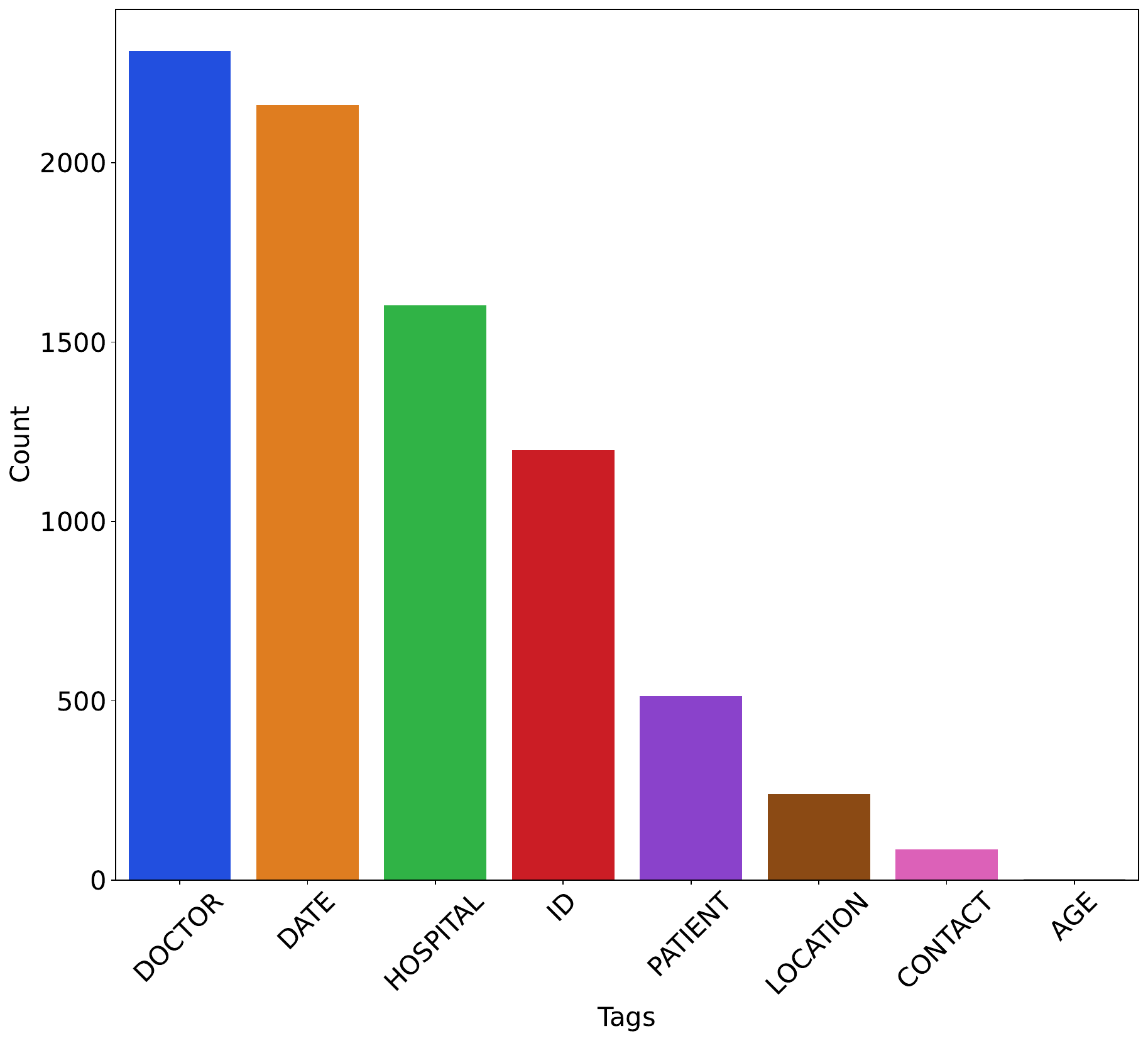}
  \caption{Tag Distribution in {\ncOld} test dataset }
  \label{fig:Tag-n2c2-2006-test}
\end{figure}

\begin{figure}[t]
  \centering
  \includegraphics[scale=0.18]{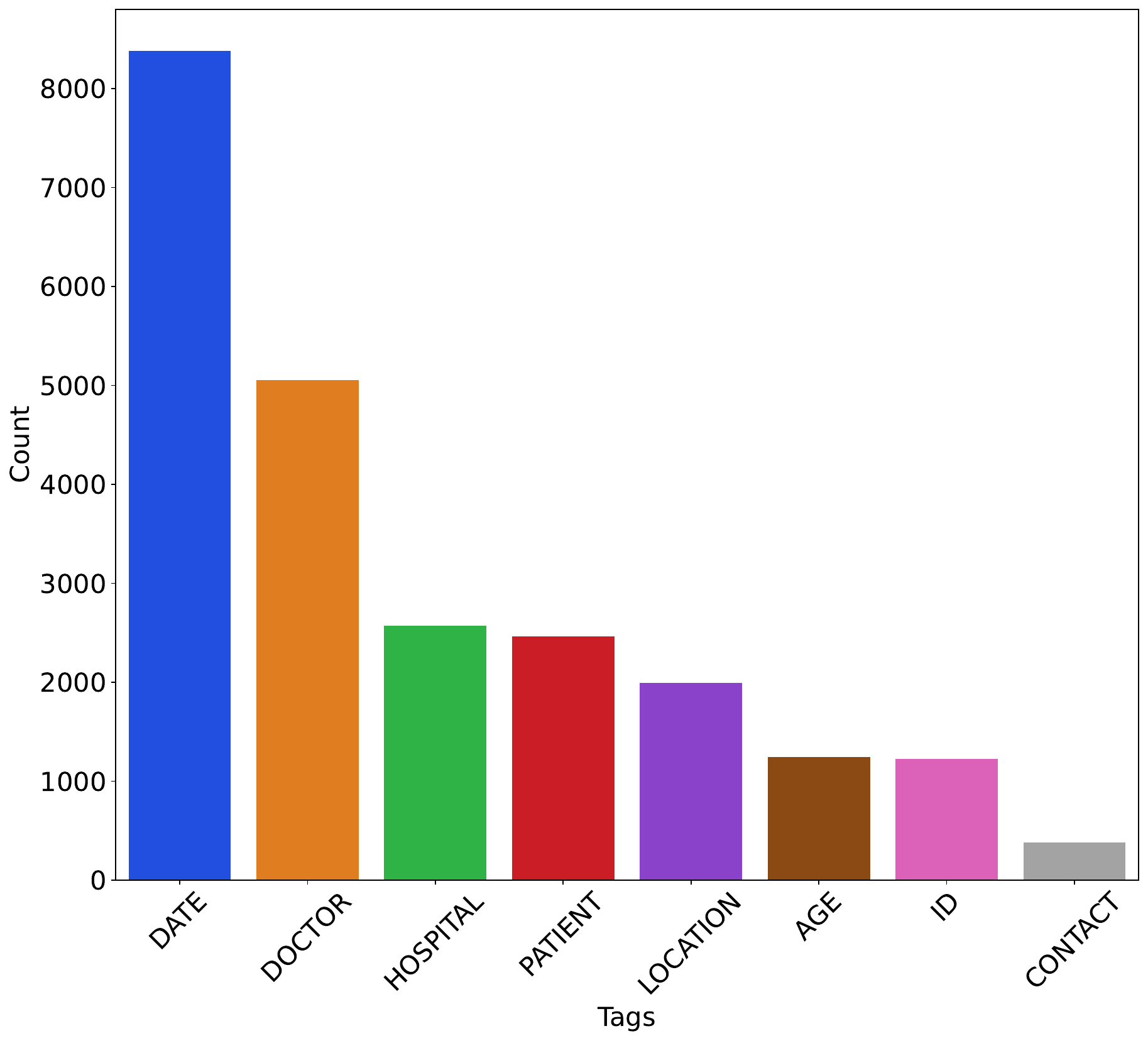}

  \caption{Tag Distribution in {\ncNew} train dataset }
  \label{fig:Tag-n2c2-2014-train}
\end{figure}

\begin{figure}[t]
  \centering
  \includegraphics[scale=0.18]{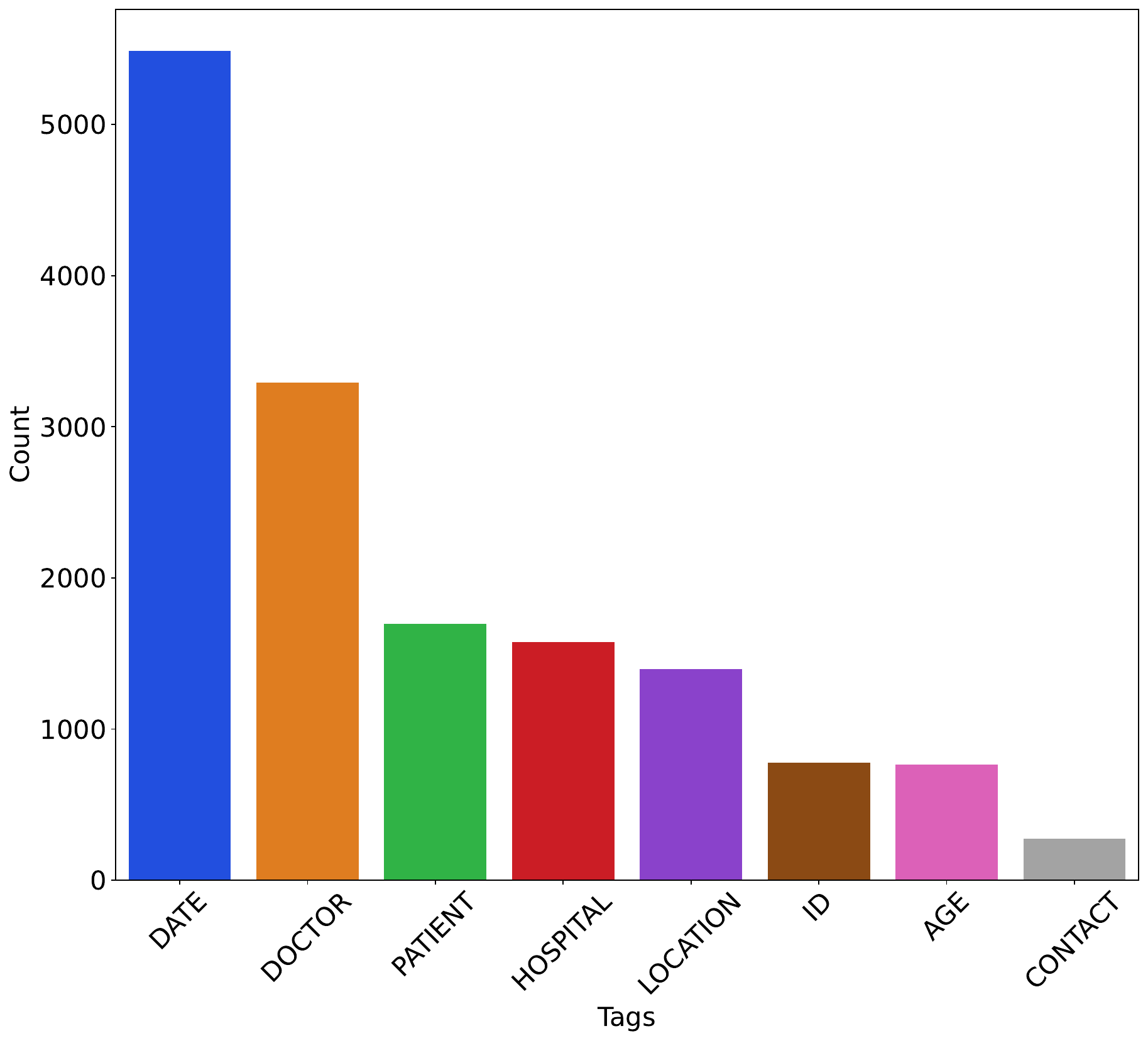}

  \caption{Tag Distribution in {\ncNew} test dataset }
  \label{fig:Tag-n2c2-2014-test}
\end{figure}

\begin{figure}[t]
  \centering
  \includegraphics[scale=0.18]{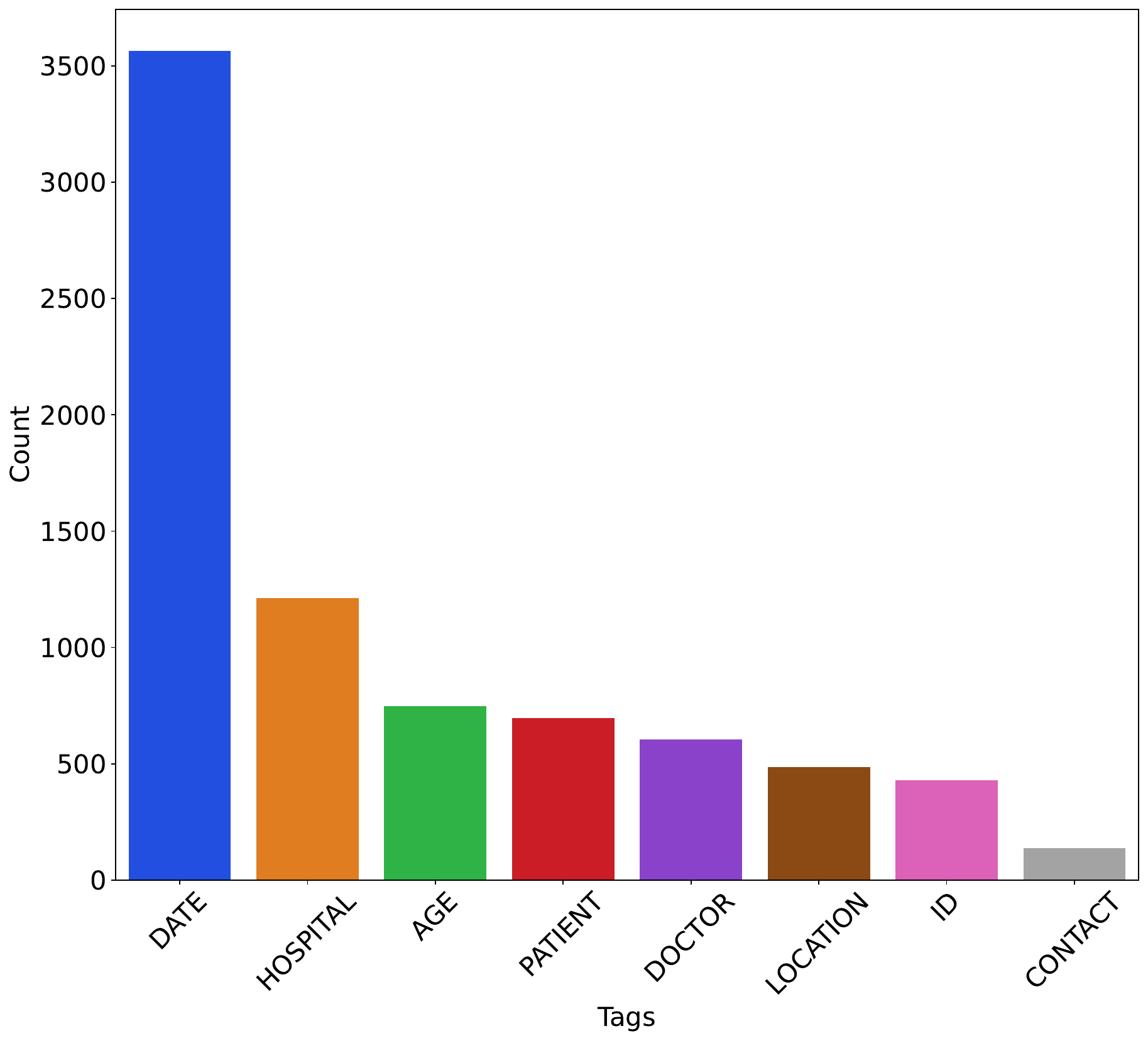}
 
  \caption{Tag distribution in {\dataReal} train dataset }
  \label{fig:Tag-ICDS-R-train}
\end{figure}

\begin{figure}[t]
  \centering
  \includegraphics[scale=0.18]{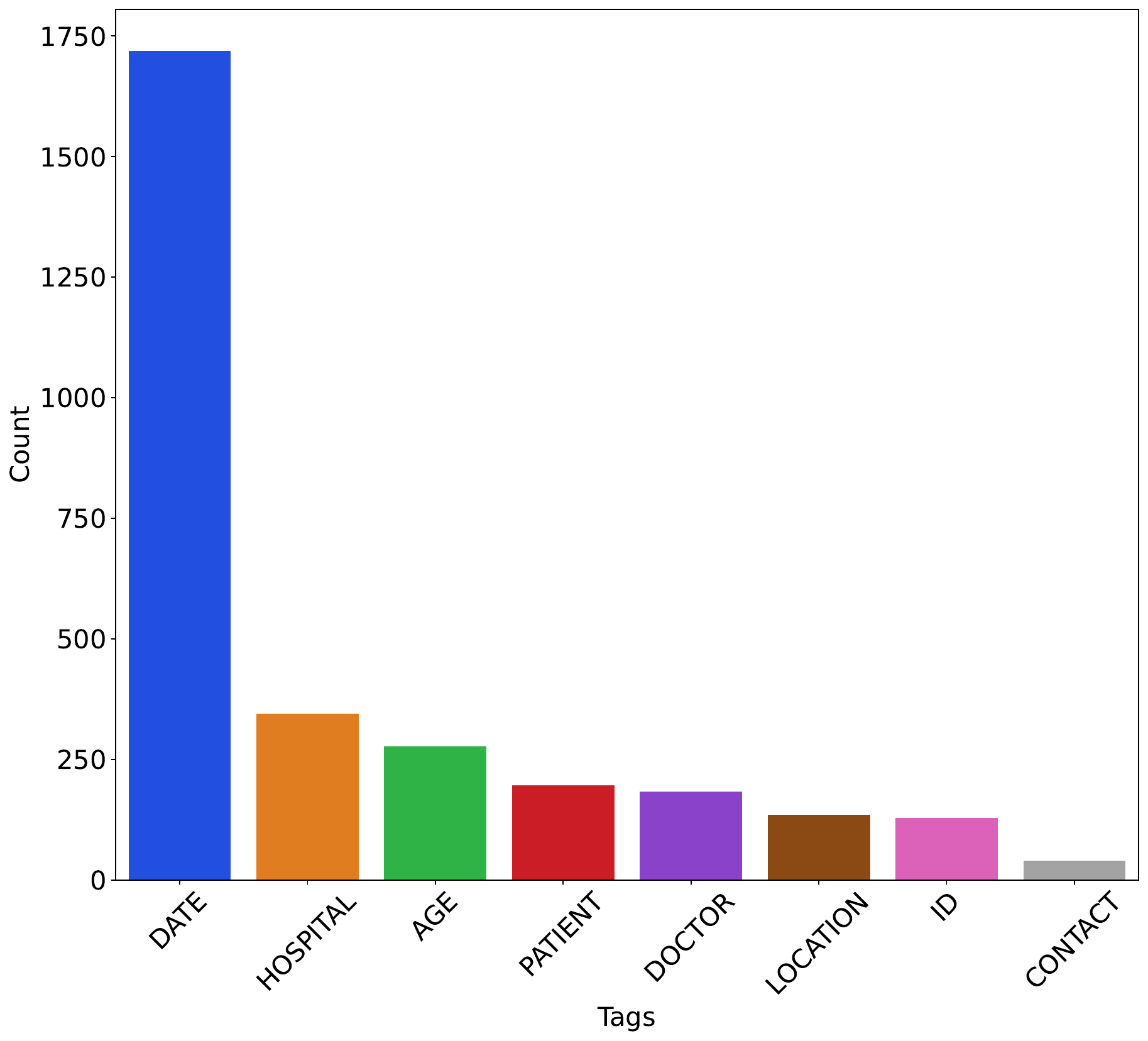}
  
  \caption{Tag distribution in {\dataReal} test dataset }
  \label{fig:Tag-ICDS-R-test}
\end{figure}

\begin{figure}[t]
  \centering
  \includegraphics[scale=0.18]{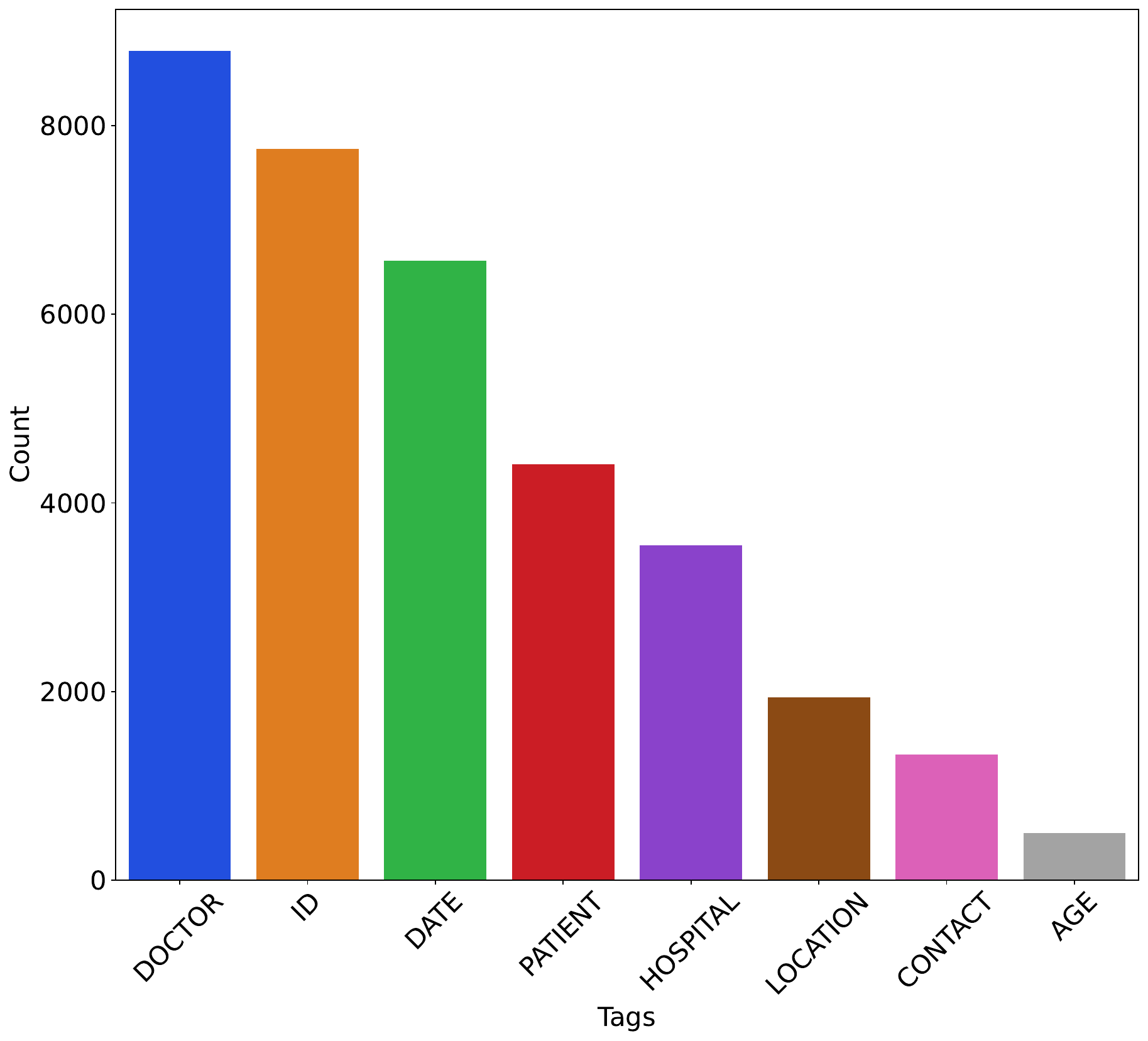}
  \caption{Tag distribution in {\dataGenGemini} dataset }
  \label{fig:Tag-ICDS-G(g)}
\end{figure}

\begin{figure}[t]
  \centering
  \includegraphics[scale=0.18]{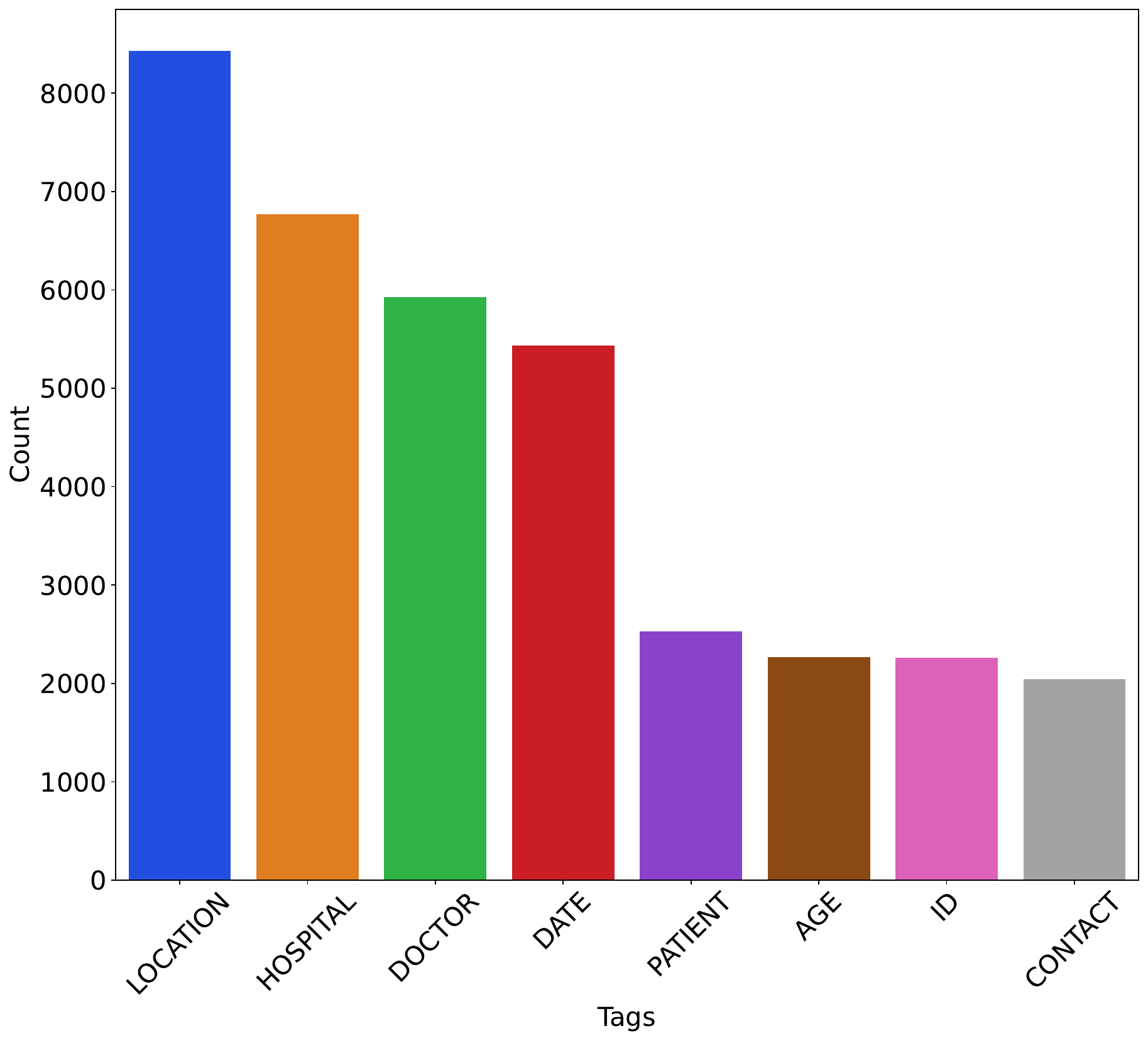}
  \caption{Tag distribution in {\dataGenLlama} dataset }
  \label{fig:Tag-ICDS-G(l)}
\end{figure}

Hence, we mapped the tag set of all the datasets to the {\ncOld} tag set. In all the datasets, we mapped entities like street, city, country, zip, etc to \texttt{LOCATION}. Similarly, we mapped phone number, mobile number, email, landline, etc, to \texttt{CONTACT}. Additionally, we mapped all the PHI-related entities to their super-set using mapping  shown in Table \ref{tab:TagMapping}. In the {\dataGenLlama} and {\dataGenGemini} datasets, we have several tags unrelated to the PHI entities. Hence, we mapped all non-PHI entities to the \texttt{OTHERS} tag. After mapping the tag set of all the datasets to {\ncOld} tag set, we calculated the tag distribution of all PHI entities across all datasets. The distribution of tag sets of all the dataset when mapped with {\ncOld} dataset are shown in Fig. \ref{fig: Tag-n2c2-2006-train}, Fig. \ref{fig:Tag-n2c2-2006-test}, Fig. \ref{fig:Tag-n2c2-2014-train}, Fig. \ref{fig:Tag-n2c2-2014-test}, Fig. \ref{fig:Tag-ICDS-R-train}, Fig. \ref{fig:Tag-ICDS-R-test}, Fig. \ref{fig:Tag-ICDS-G(g)}, and Fig. \ref{fig:Tag-ICDS-G(l)}.



\section{Corpus Statistics}

\begin{figure}[t]
  \centering
  \includegraphics[scale=0.18]{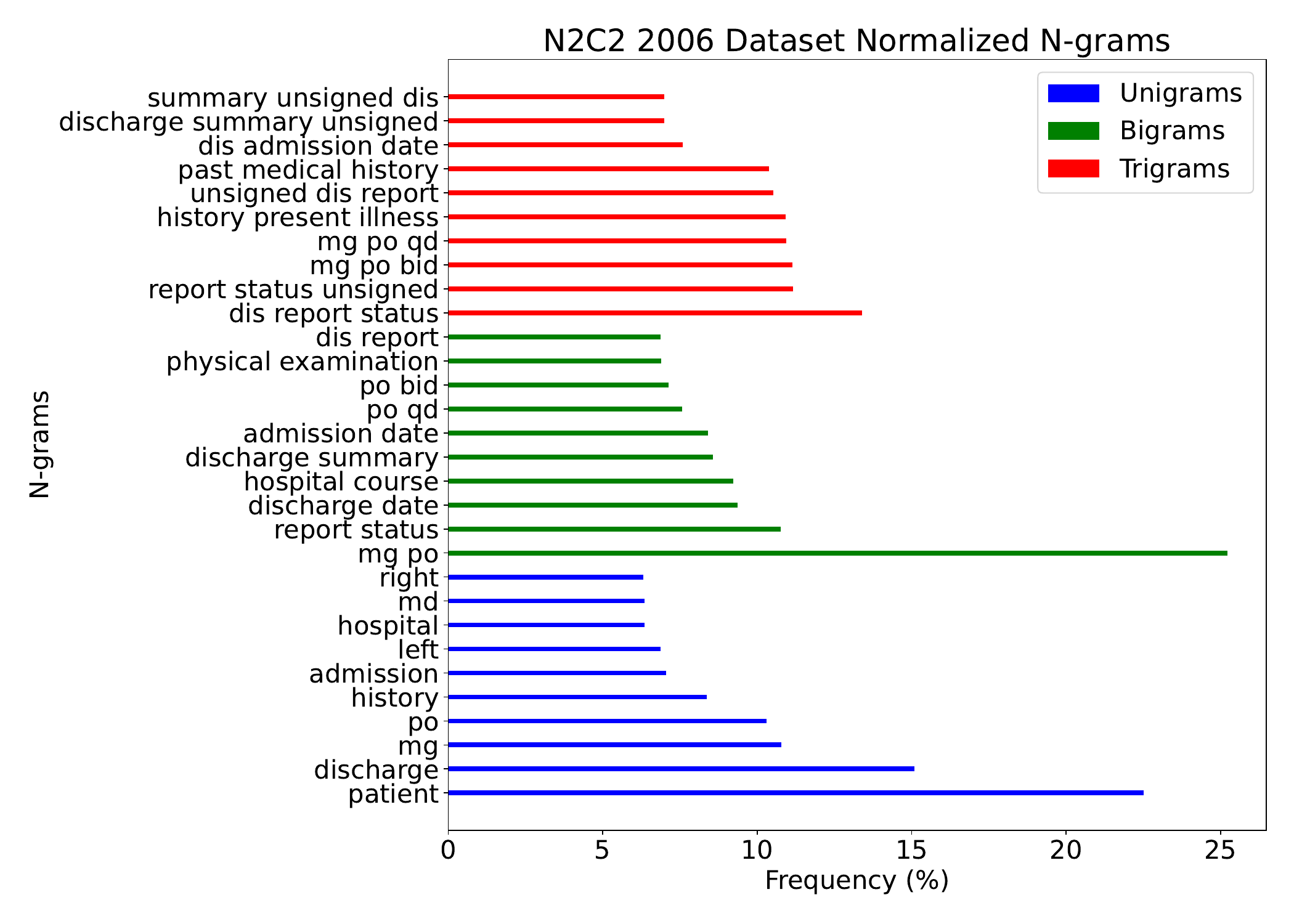}
  \caption{n2c2-2006 Top 10 N-grams}
  \label{fig:n2c2realngrams}
\end{figure}

\begin{figure}[t]
  \centering
  \includegraphics[scale=0.18]{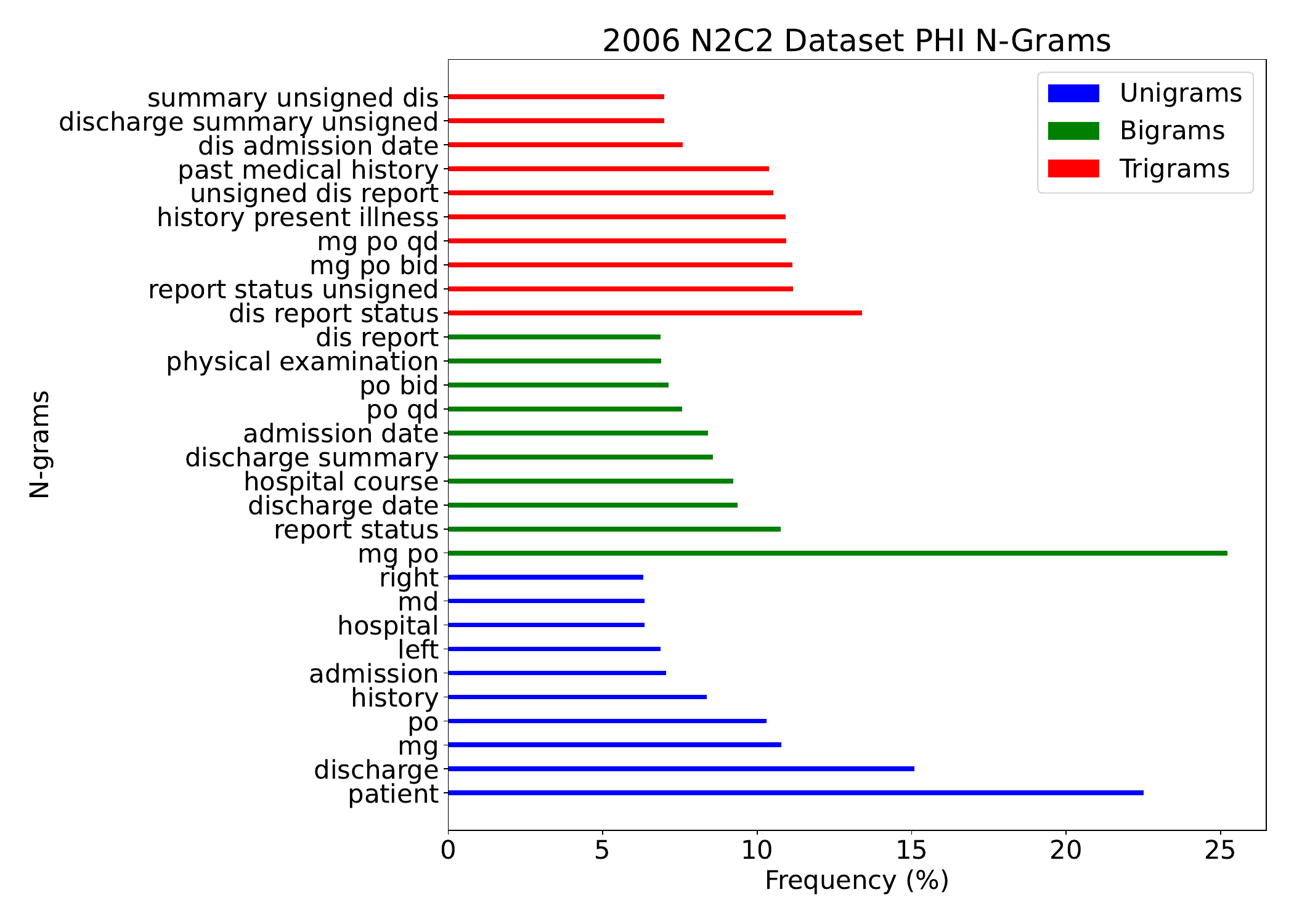}
  \caption{n2c2-2006 Top 10 N-grams Spanning PHI Elements}
  \label{fig:PHIn2c2ngrams}
\end{figure}

\begin{figure}[t]
  \centering
  \includegraphics[scale=0.18]{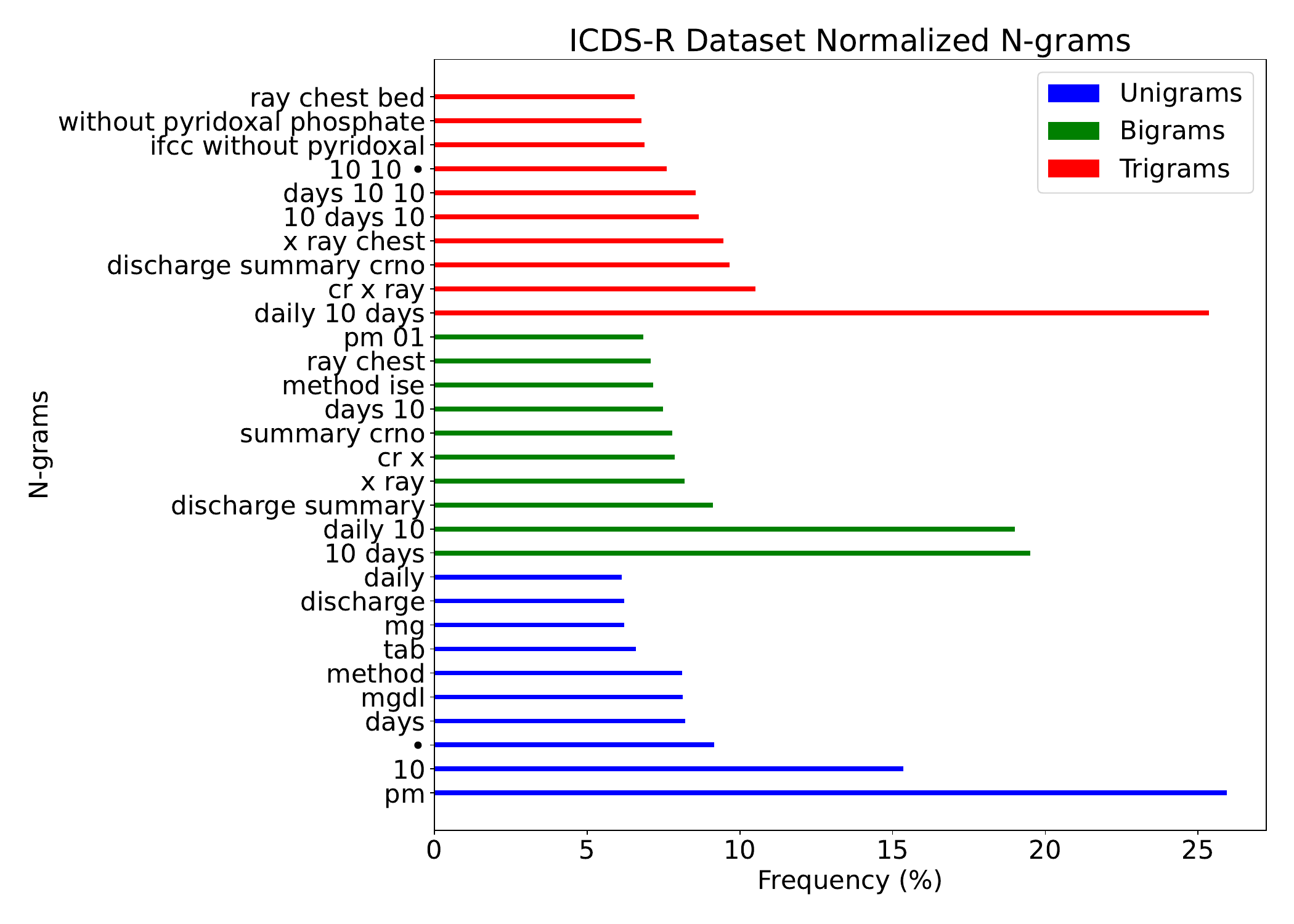}
  \caption{{\dataReal} Top 10 N-grams}
  \label{fig:sgpgirealngrams}
\end{figure}

\begin{figure}[t]
  \centering
  \includegraphics[scale=0.18]{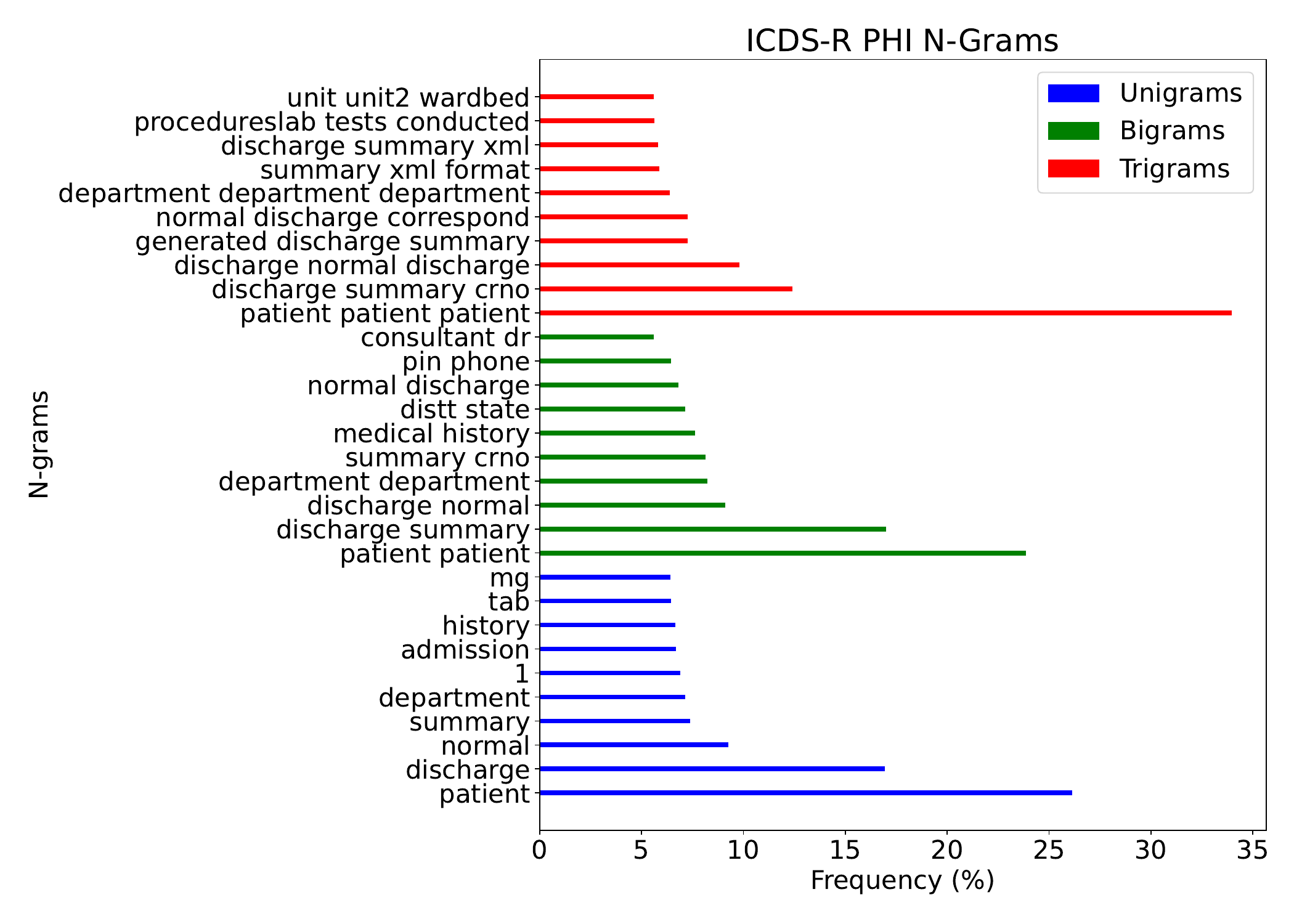}
  \caption{{\dataReal} Top 10 N-grams Spanning PHI Elements}
  \label{fig:PHIsgpgirealngrams}
\end{figure}

The n-gram frequencies from the {\ncOld} dataset show a strong emphasis on clinical and procedural language, including terms like `mg,' `po,' and `hospital,' as shown in Fig. \ref{fig:n2c2realngrams}. Notably, phrases such as `discharge summary' and `physical examination' dominate, highlighting standard documentation practices. Trigrams such as `dis report status' and `report status unsigned' indicate typical phrasing in medical reports. This is in contrast with the {\dataReal} dataset in Fig. \ref{fig:sgpgirealngrams}, where there is a predominance of time-related unigrams (`pm,' `days') and clinical terms (`mgdl,' `method'). The frequent bigrams and trigrams revolve around treatment and diagnosis descriptors, such as `daily 10 days' and `x ray chest,' illustrating the detailed recording of patient care routines and diagnostic procedures commonly found in medical records. In the {\ncOld} dataset, bigrams like `mg po' and `discharge date,' and trigrams like `mg po bid' and `history present illness,' which reveal specific medication dosages and detailed descriptions of patient conditions, are found next to PHI elements, as shown in Fig. \ref{fig:PHIn2c2ngrams}. In the {\dataReal} dataset, specific trigrams like `discharge summary crno' and `normal discharge correspond' are located near PHI elements (Fig. \ref{fig:PHIsgpgirealngrams}). The differences between the n2c2 2006 dataset and {\dataReal} highlight how clinical documentation practices and language differ between the US and India.

\begin{figure}[t]
  \centering
  \includegraphics[scale=0.18]{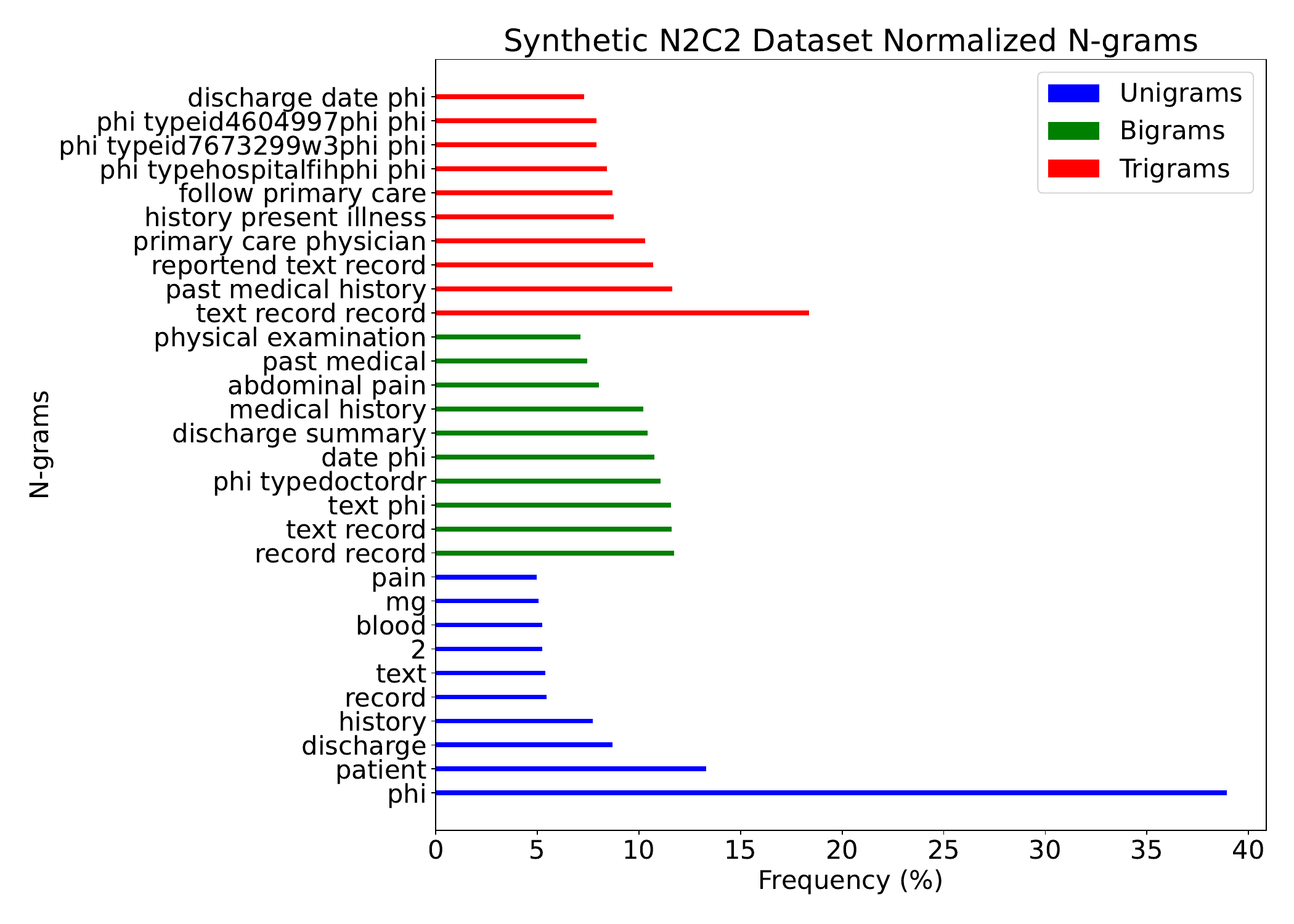}
  \caption{{\dataGenGemini} Top 10 N-grams}
  \label{fig:n2c2synngrams}
\end{figure}

\begin{figure}[t]
  \centering
  \includegraphics[scale=0.18]{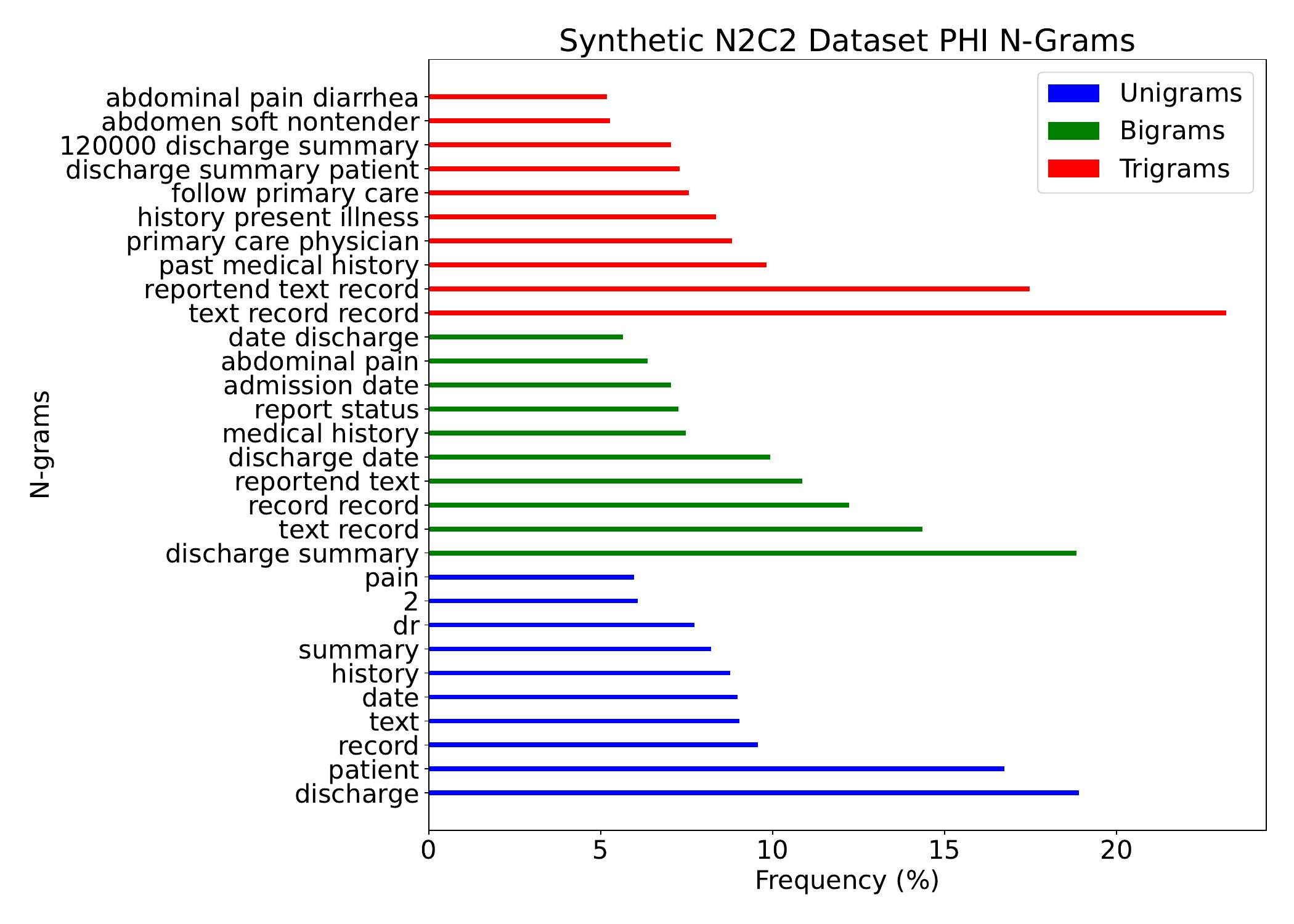}
  \caption{{\dataGenGemini} Top 10 N-grams Spanning PHI Elements}
  \label{fig:PHIn2c2synngrams}
\end{figure}

\begin{figure}[t]
  \centering
  \includegraphics[scale=0.18]{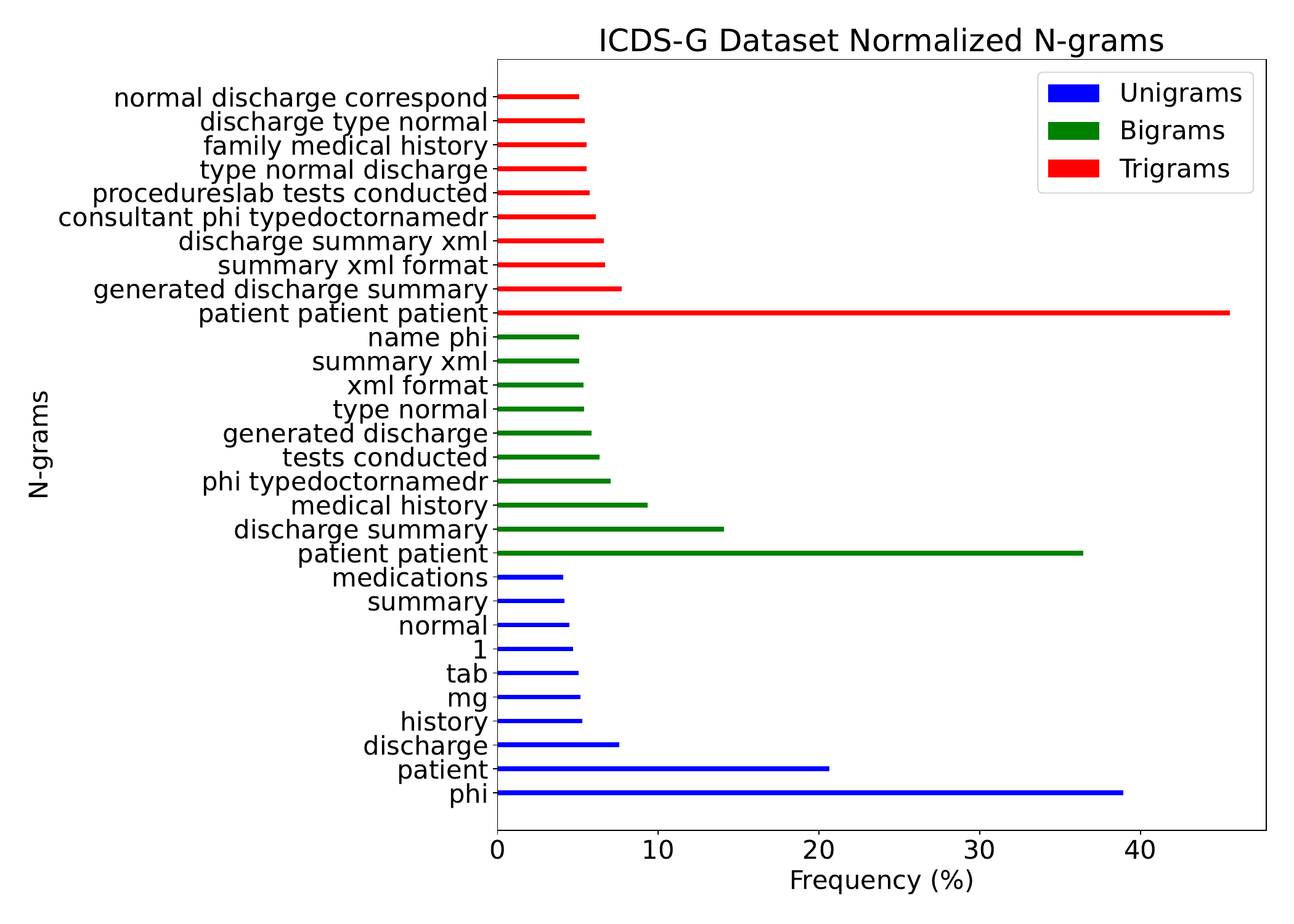}
  \caption{{\dataGenLlama} Top 10 N-grams}
  \label{fig:sgpgisynngrams}
\end{figure}

\begin{figure}[t]
  \centering
  \includegraphics[scale=0.18]{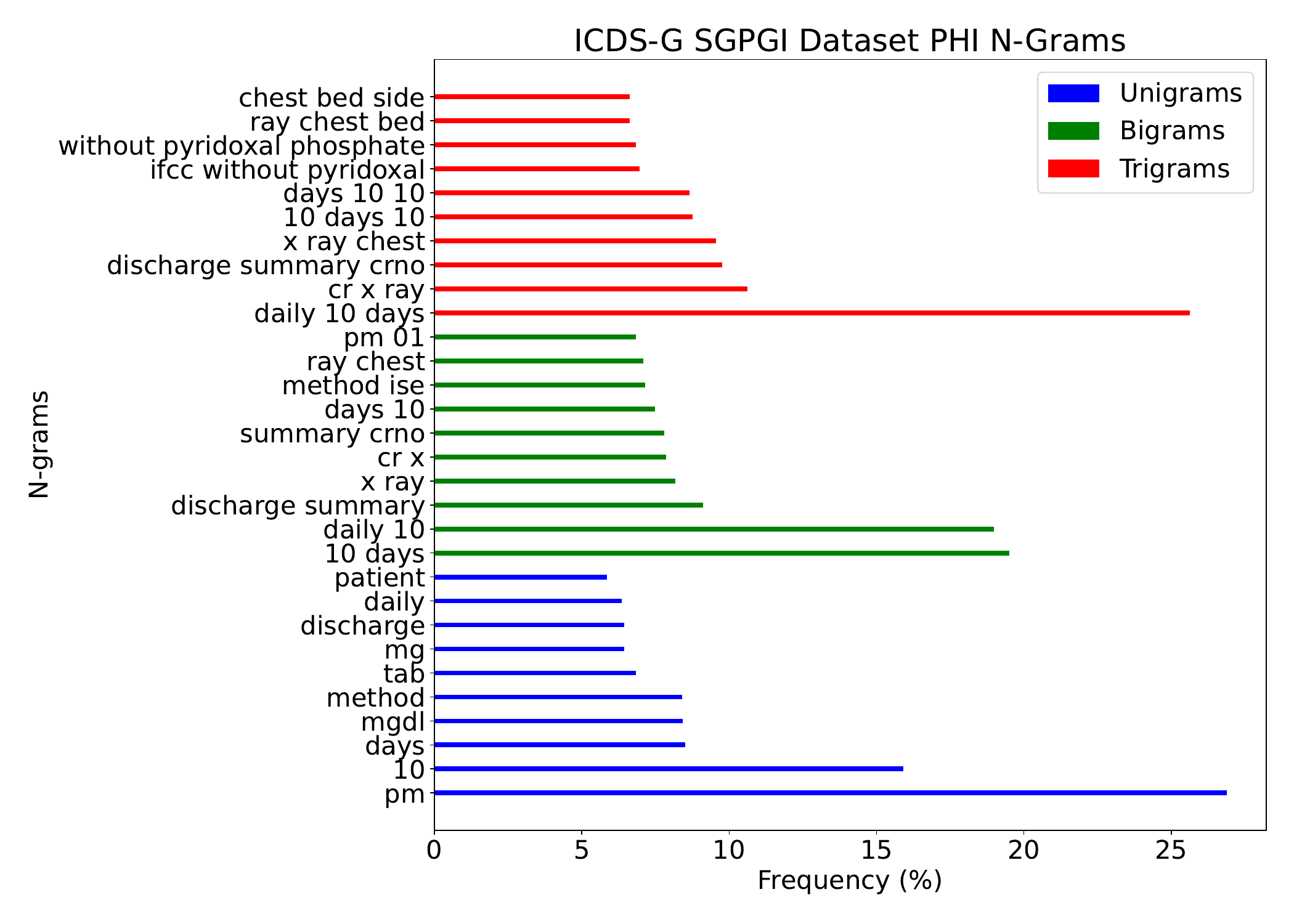}
  \caption{{\dataGenLlama} Top 10 N-grams Spanning PHI Elements}
  \label{fig:PHIsgpgisynngrams}
\end{figure}

In the synthetic {\dataGenGemini} dataset, the frequent occurrence of `phi' in various n-grams highlights (in Fig. \ref{fig:n2c2synngrams}) the inclusion of potentially identifiable information. Trigrams such as `phi typehospitalfihphi' and `phi typeid7673299w3phi' illustrate the use of placeholders for personal identifiers, indicative of the synthetic nature of the dataset and its focus on mimicking real-world PHI data while maintaining privacy. In the {\dataGenLlama} dataset, the frequent mention of basic terms like `patient,' `discharge,' and `history' reflects their regular usage in clinical documents, as seen in Fig. \ref{fig:sgpgisynngrams}. Phrases such as `discharge summary' and `medical history' indicate standardized document formats. For n-grams next to PHI elements in the synthetic {\dataGenGemini} dataset as seen in Fig. \ref{fig:PHIn2c2synngrams}, we observe a mix of clinical terminology (`discharge,' `patient,' `history') and documentation descriptors (`text record,' `reportend text'). Bigrams and trigrams like `discharge summary patient' and `text record record' suggest a replication of typical medical documentation formats. Terms like `primary care physician' and `history present illness' reflect the comprehensive nature of clinical narratives. In contrast, the n-grams next to PHI elements in the {\dataGenLlama} dataset, as shown in Fig. \ref{fig:PHIsgpgisynngrams}, highlight the frequent use of both temporal (`pm', `days') and medical (`mgdl,' `discharge') terms. Common bigrams and trigrams such as `discharge summary,' `cr x ray,' and `x ray chest' underscore the clinical focus on diagnostic imaging and summary documentation. The trigrams involving `daily 10 days' and `x ray chest bed' reflect specific medical interventions and patient care protocols typically documented in patient records.
\begin{figure}[!b]
  \centering
  \includegraphics[scale=0.30]{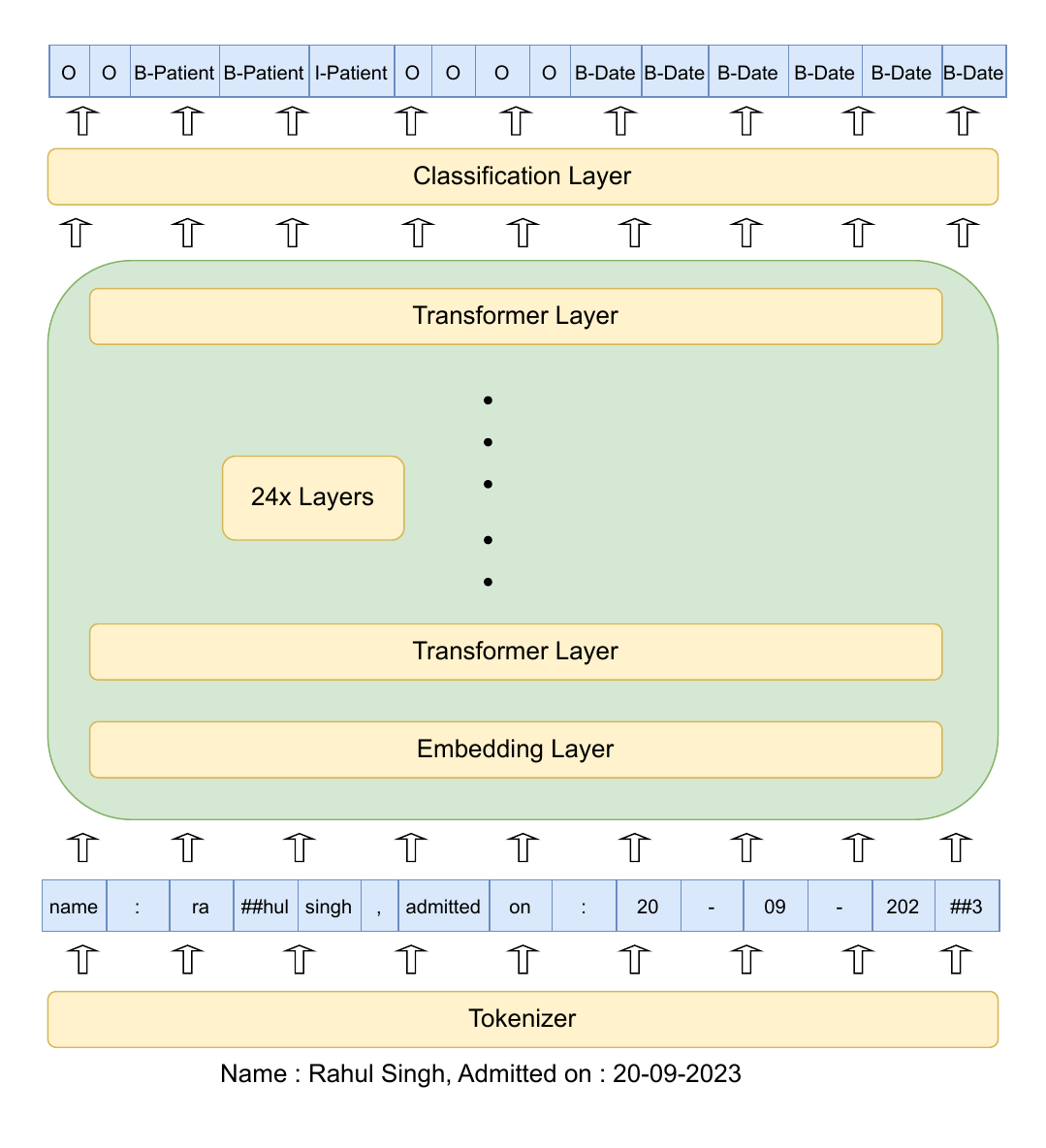}
  \caption{Architecture of \model }
  \label{fig:model-architecture}
\end{figure}

\begin{figure}[t]
  \centering
  \includegraphics[scale=0.18]{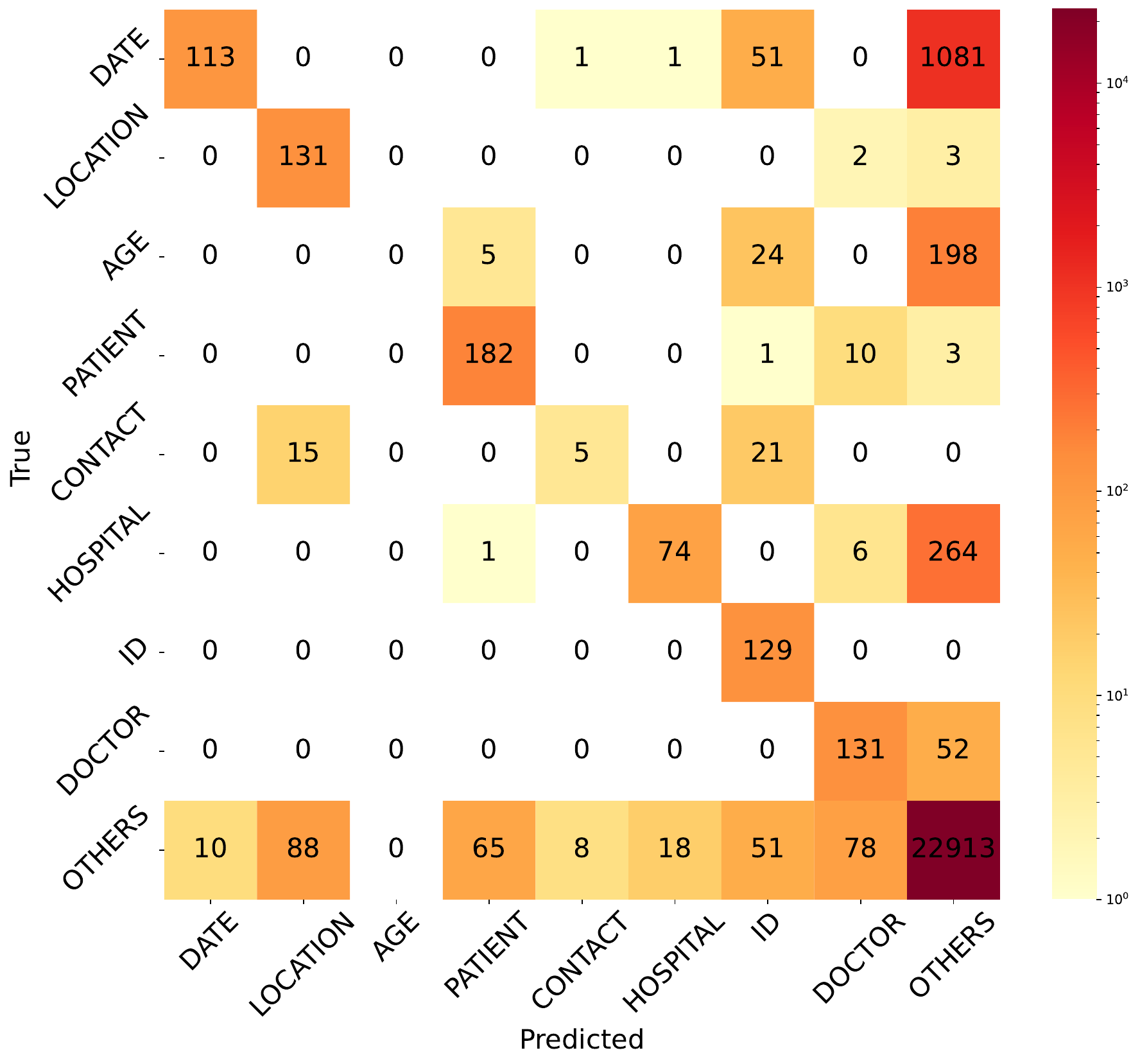}
  \small
  \caption{Confusion matrix on {\dataReal} test set when {\model} finetuned on {\ncOld} }
  \label{fig:cm_2006}
\end{figure}

\begin{figure}[t]
  \centering
  \includegraphics[scale=0.18]{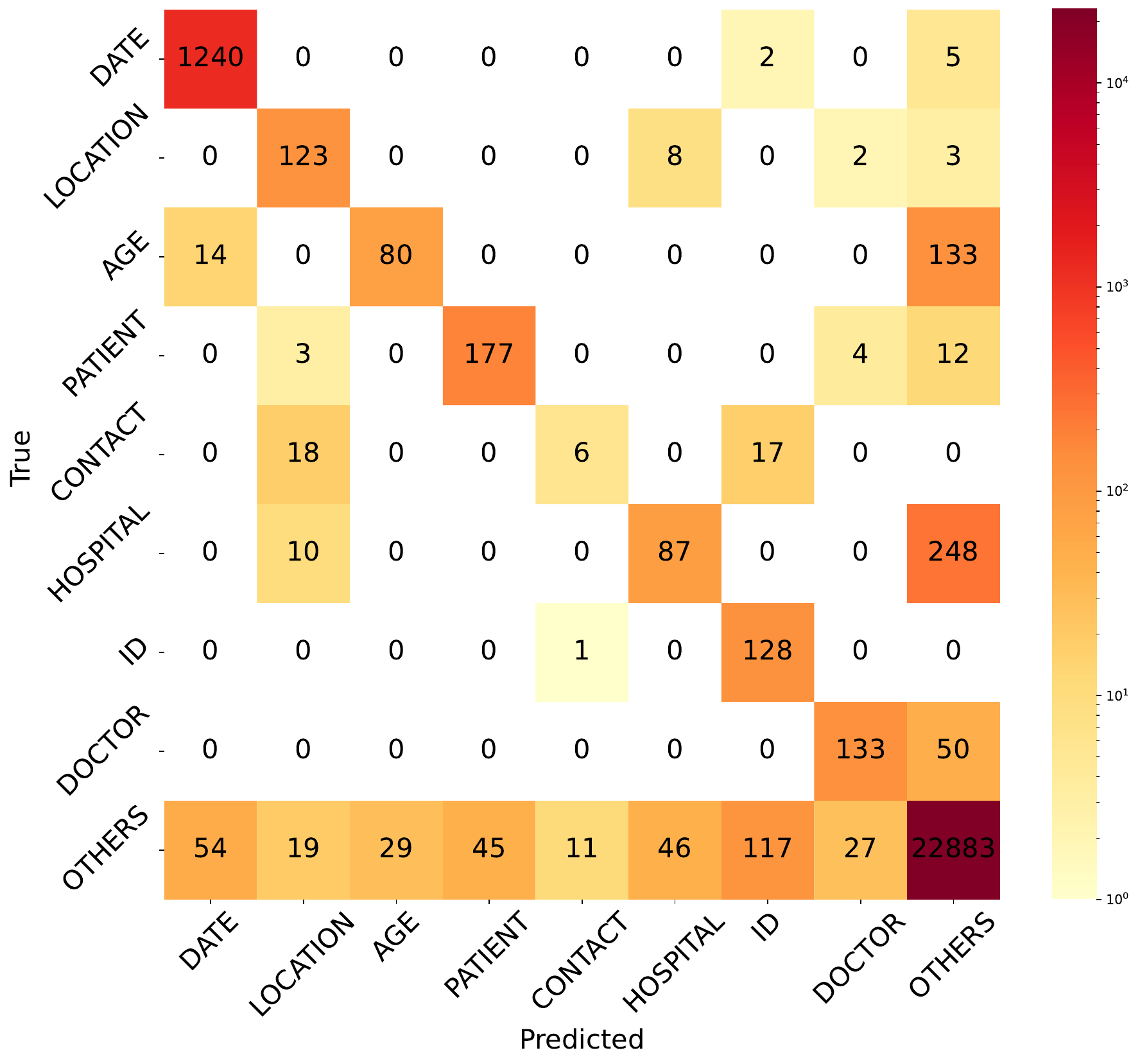}
  \small
  \caption{ Confusion matrix on {\dataReal} test set when {\model} finetuned on {\ncNew} }
  \label{fig:cm_2014}
\end{figure}

\begin{figure}[t]
  \centering
  \includegraphics[scale=0.18]{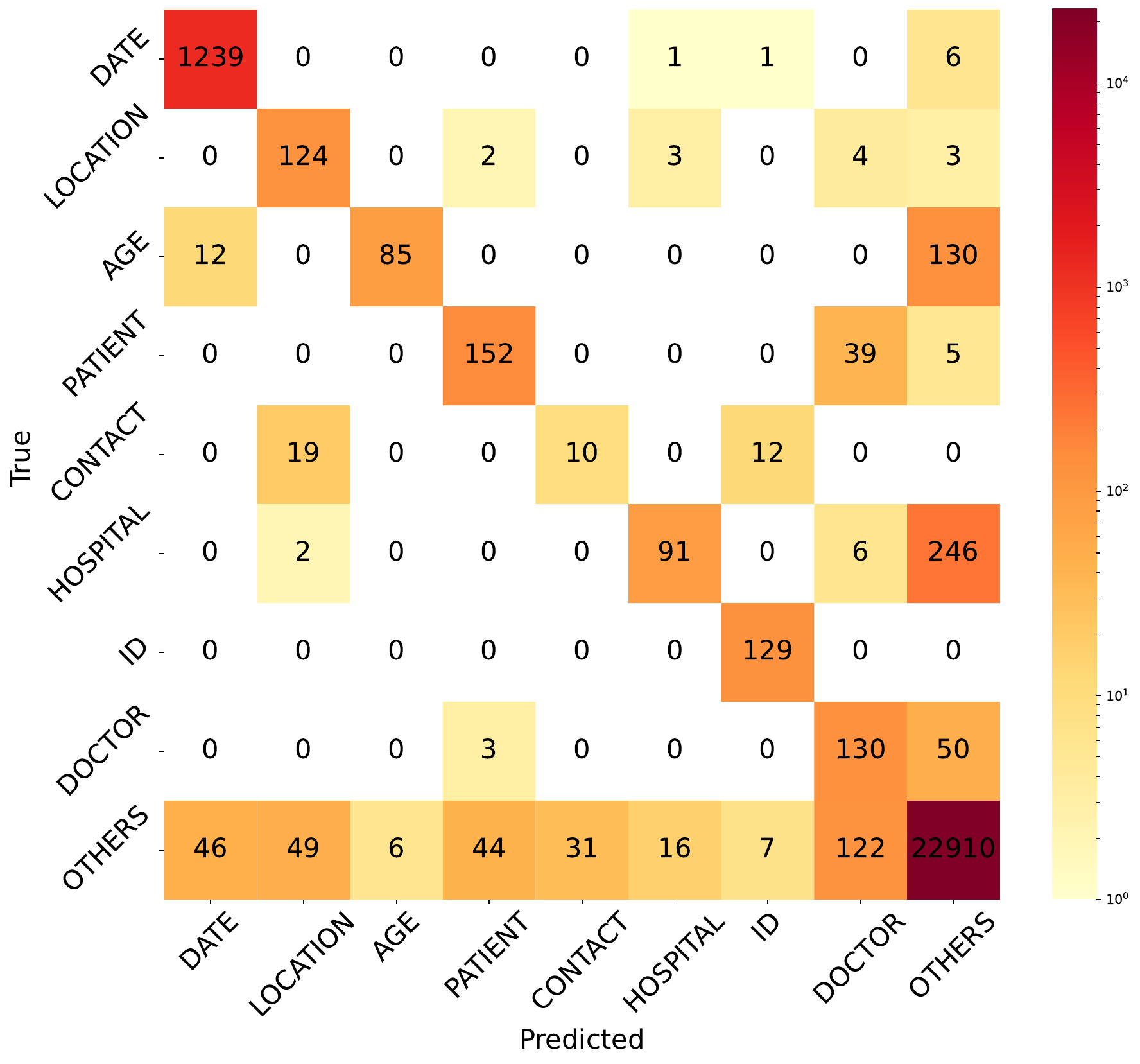}
  \small
  \caption{Confusion matrix on {\dataReal} test set when {\model} finetuned on Combining {\ncOld} and {\ncNew} }
  \label{fig:cm_2006+2014}
\end{figure}

\begin{figure}[t]
  \centering
  \includegraphics[scale=0.18]{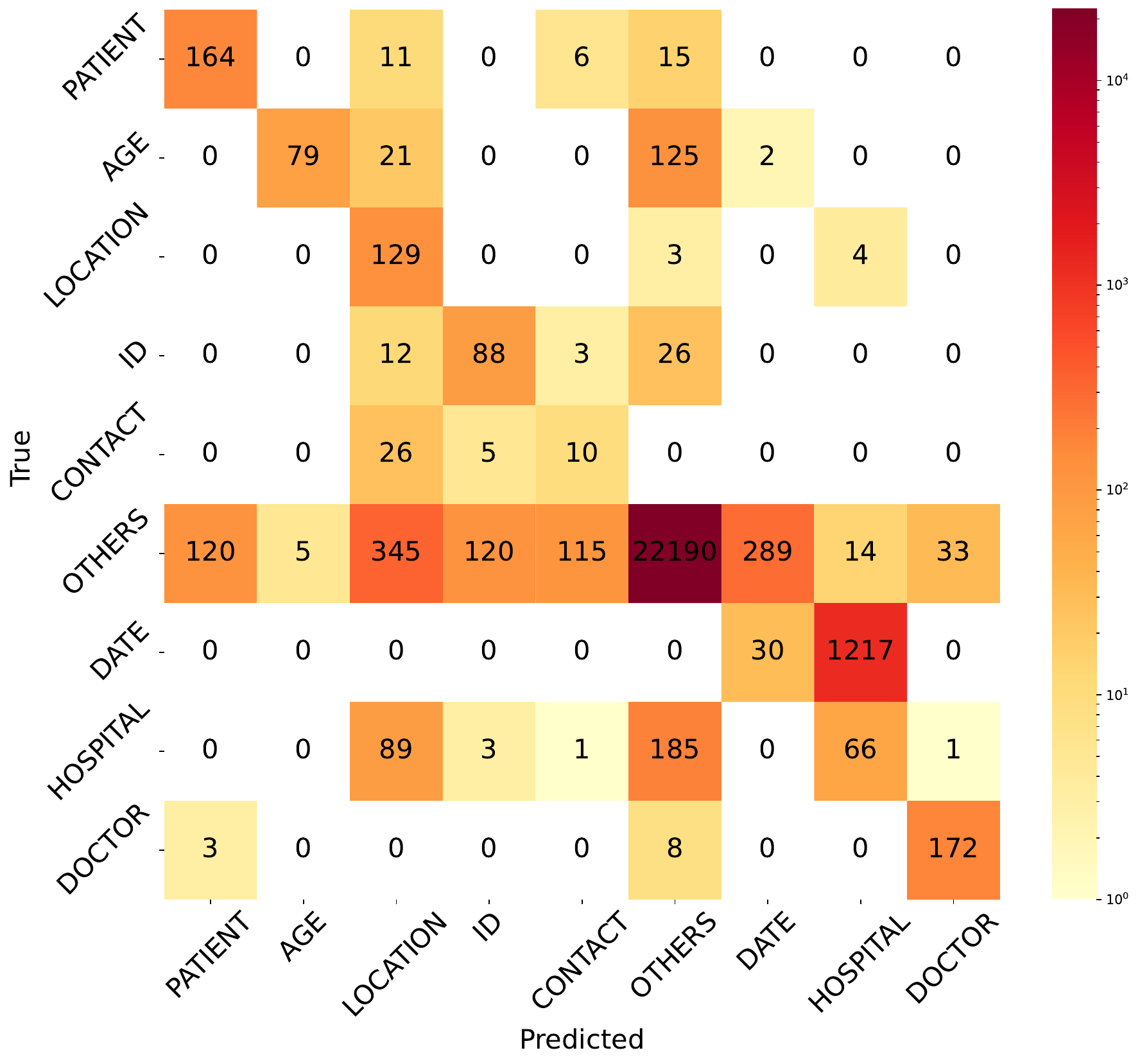}
  \small
  \caption{Confusion matrix on {\dataReal} test set when {\model} finetuned on {\dataGenGemini} }
  \label{fig:cm_ICDS-G-g}
\end{figure}

\begin{figure}[t]
  \centering
  \includegraphics[scale=0.18]{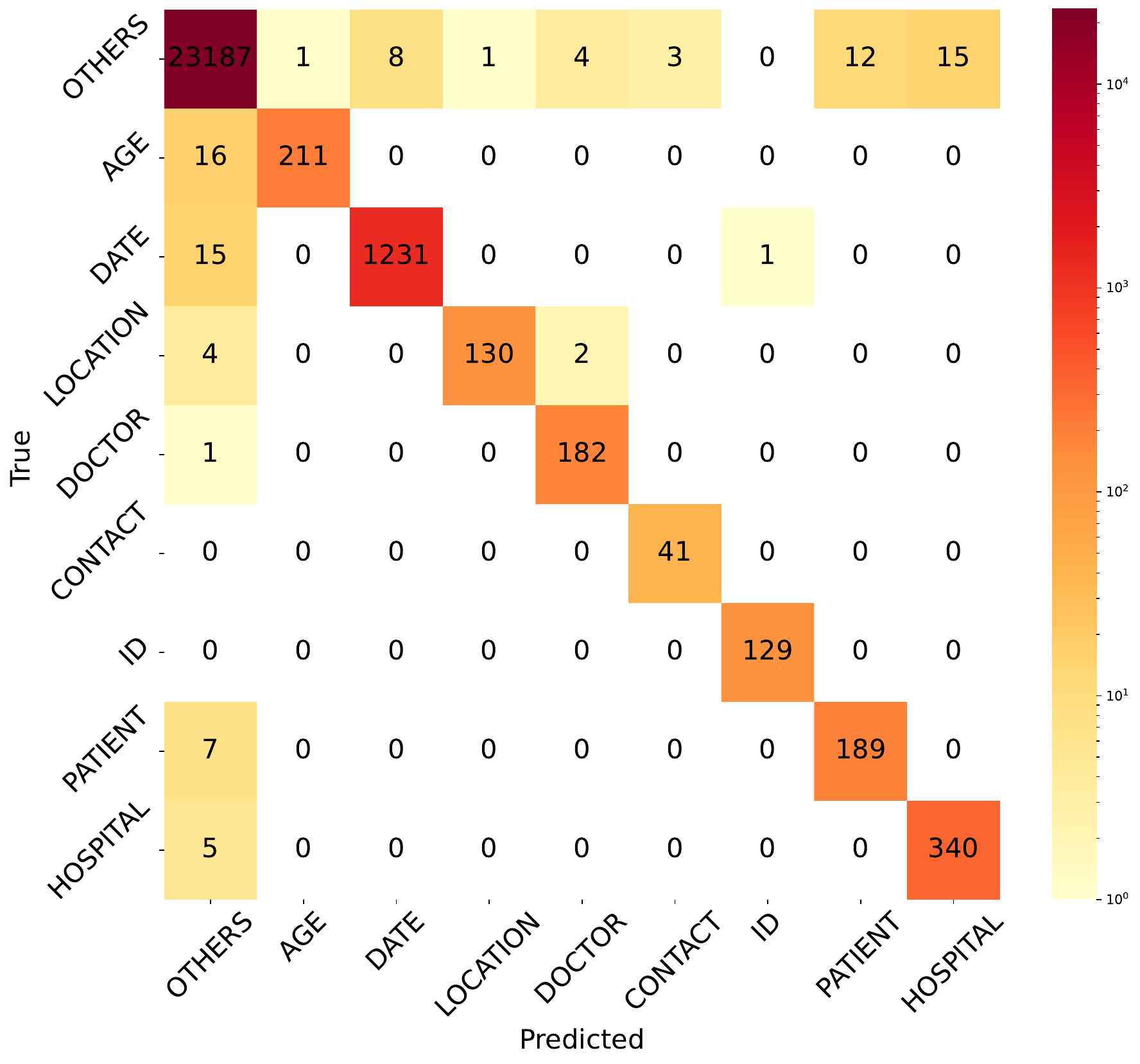}
  \caption{Confusion matrix on {\dataReal} test set when {\model} finetuned on {\dataGenLlama} }
  \label{fig:cm_ICSD-G-l}
\end{figure}

\begin{figure}[t]
  \centering
  \includegraphics[scale=0.18]{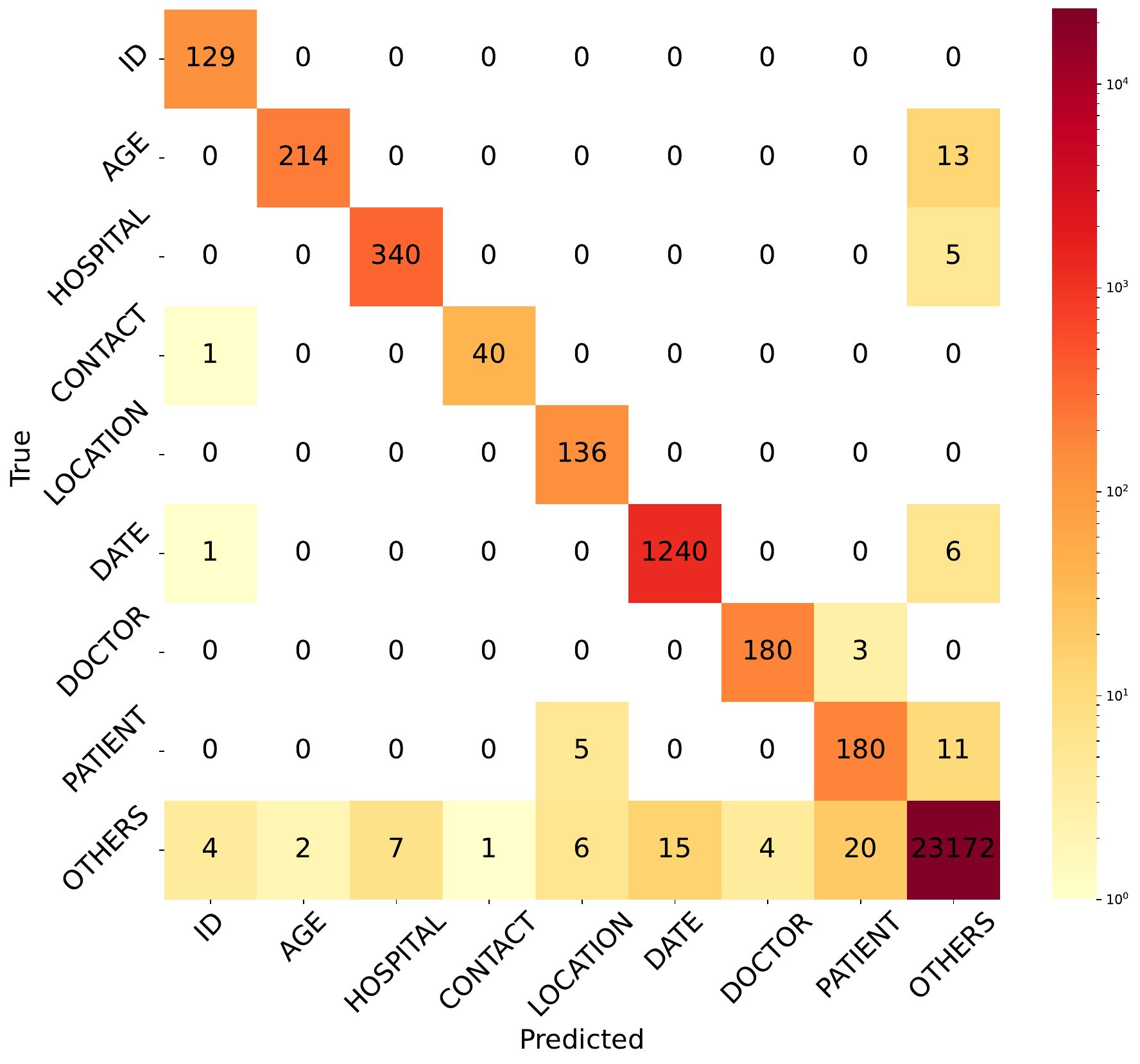}
  \caption{Confusion matrix on {\dataReal} test set when {\model} finetuned on Combining {\dataGenLlama} and {\dataGenGemini} dataset }
  \label{fig:cm_ICSD-G-l+ICSD-G-g}
\end{figure}


\begin{figure*}[t]
  \centering
  \begin{subfigure}[b]{0.48\textwidth}
    \centering
    \includegraphics[scale=0.18]{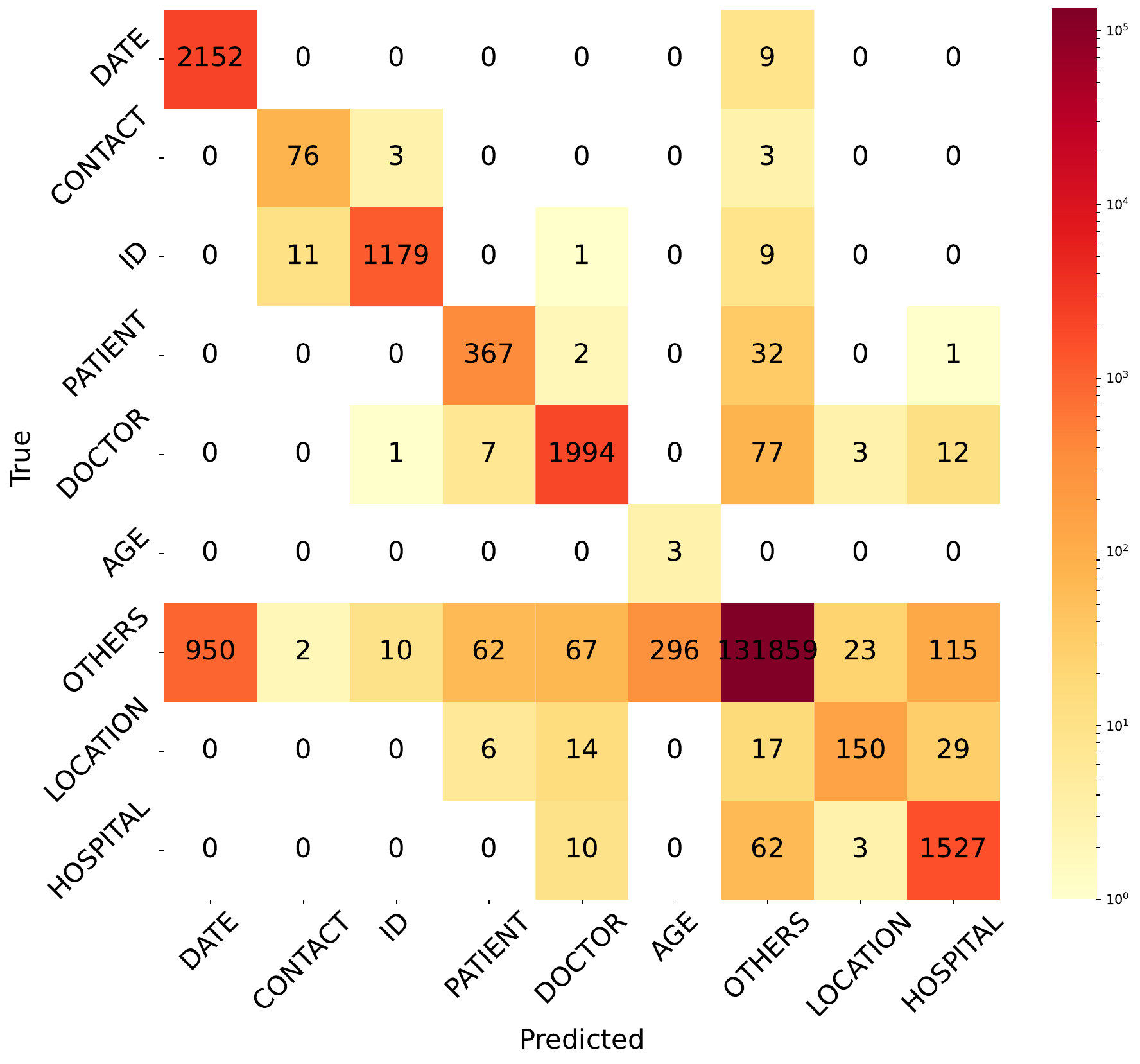}
    \caption{Confusion matrix on {\ncOld} test set when {\model} finetuned on combining {\ncNew}, {\dataGenGemini}, and {\dataGenLlama} dataset.}
    \label{fig:cm_n2c2_2014+ICSD-G-g+ICSD-G-l}
  \end{subfigure}
  \hfill
  \begin{subfigure}[b]{0.48\textwidth}
    \centering
    \includegraphics[scale=0.18]{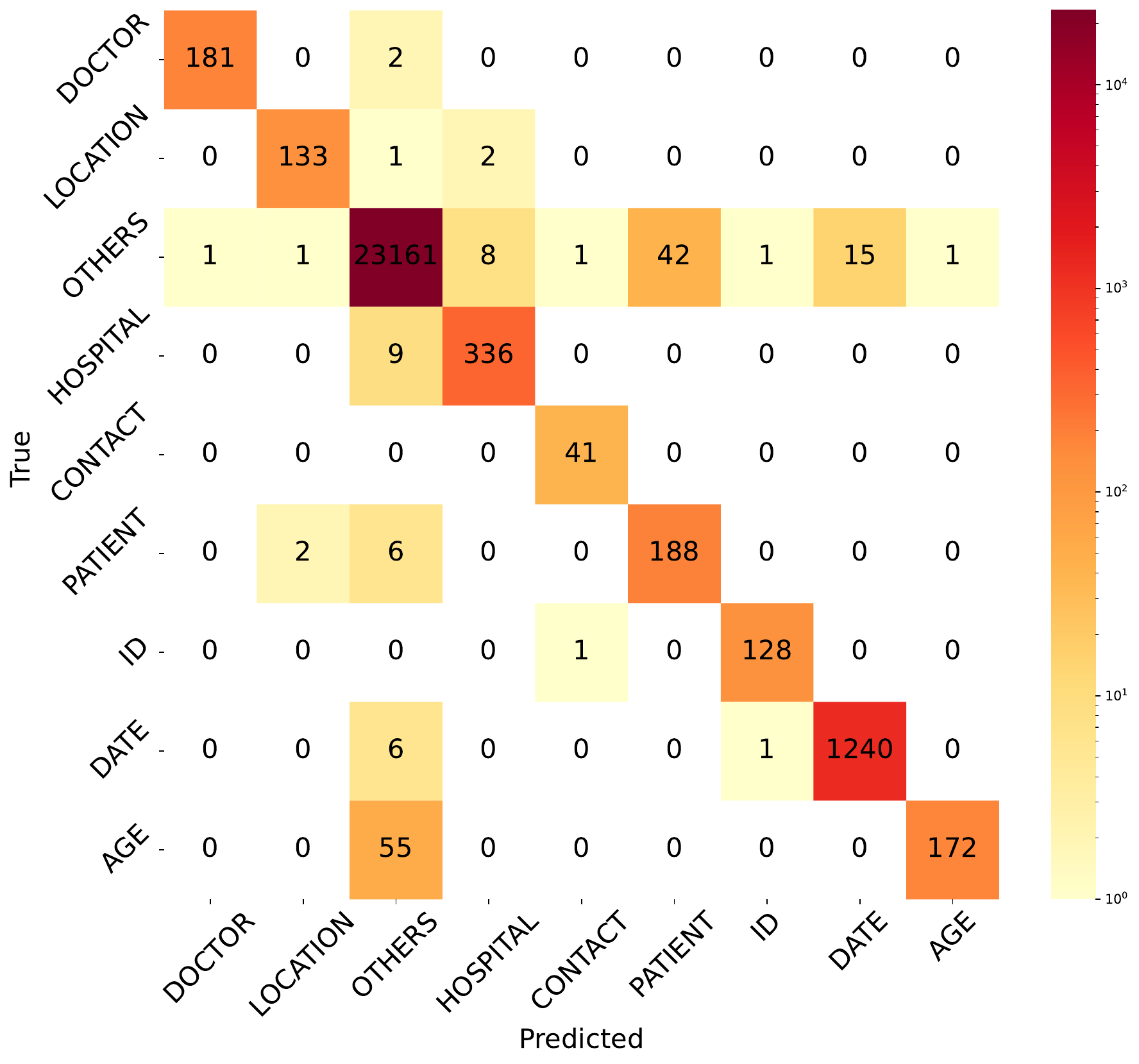}
    \caption{Confusion matrix on {\dataReal} test set when {\model} finetuned on combining {\dataGenLlama}, {\dataGenGemini}, {\ncOld}, and {\ncNew} dataset.}
    \label{fig:cm_n2c2_2014+n2c2_2006+ICSD-G-g+ICSD-G-l}
  \end{subfigure}
  \caption{Confusion matrix on {\ncOld} and {\dataReal} testset when {\model} fine-tuned on combination of generated and real data.}
  \label{fig:combined_confusion_matrices}
\end{figure*}

\section{Model Training Details} \label{app:model-training}

We fine-tuned dslim/bert-base-NER \cite{huggingfaceDslimbertbaseNERHugging}, ghadeermobasher/BCHEM4-Modified-BioBERT-v1 \cite{huggingfaceGhadeermobasherBCHEM4ModifiedBioBERTv1Hugging}, and Clinical-AI-Apollo/Medical-NER \cite{huggingfaceClinicalAIApolloMedicalNERHugging}. We obtained a consistent train-set F1 Score for PHI entities from these models after fine-tuning, but the performance of these models decreased significantly when we tested them on cross-dataset settings. However, after fine-tuning, {\model}  outperformed these models in the same and cross-dataset settings, so we chose \model\ for further experiments. Fig. \ref{fig:model-architecture} shows the model architecture.

\noindent\model\ was fine-tuned on each training set as given in Table \ref{tab:experiments-matrix} and tested on each corresponding test set. We fixed the hyperparameters for all the experiments. The model was fine-tuned at four epochs in all the experiments with a batch size of 8; the learning rate was 5e-5.  We used Weighted Cross entropy loss to handle the data imbalance problem because around 90 percent of the tokens correspond to non-PHI entities in all datasets. After several experiments, we devised a formula to assign weights to different Entities. 
$w_t = \log \left( 4 \times \frac{n}{n_t} \right), $ where 
\(w_t \)  is the weight assigned to the  \(t^{th} \)entity; 
\(n_t\) is the number of tokens in the  \(t^{th} \) entity; 
\(n \) is the total number of tokens in the dataset

\section{Evaluation Metrics} \label{app:metrics}
Model was evaluated using various performance metrics as described below.

\begin{itemize}
    \item Macro Precision: 
    \begin{equation*}
    \small
    \text{Precision}_\text{macro} =
    \frac{1}{n}  \sum_{i=1}^{n}  \frac{TP_{i}}{TP_{i}+FP_{i}}
\end{equation*}

\item Macro Recall 
\begin{equation*}
\small
   \text{Recall}_\text{macro} =
    \frac{1}{n}  \sum_{i=1}^{n}  \frac{TP_{i}}{TP_{i}+FN_{i}}
\end{equation*}

\item Macro F1-score
\begin{equation*}
\small
   \text{F1-score}_\text{macro} = \frac{2 \times \text{Precision}_\text{macro}\times \text{Recall}_\text{macro}}{\text{Precision}_\text{macro}+\text{Recall}_\text{macro}}
\end{equation*}

    \item Micro Precision: 
    \begin{equation*}
    \small
    \text{Precision}_\text{micro} =
    \frac{\sum_{i=1}^{n} TP_{i}}{\sum_{i=1}^{n} (TP_{i} + FP_{i})}
\end{equation*}

\item Micro Recall 
\begin{equation*}
\small
   \text{Recall}_\text{micro} =
    \frac{\sum_{i=1}^{n} TP_{i}}{\sum_{i=1}^{n} (TP_{i} + FN_{i})}
\end{equation*}

\item Micro F1-score
\begin{equation*}
\small
   \text{F1-score}_\text{micro} = \frac{2 \times \text{Precision}_\text{micro}\times \text{Recall}_\text{micro}}{\text{Precision}_\text{micro}+\text{Recall}_\text{micro}}
\end{equation*}

\end{itemize}

\end{document}